%% file: counterfactual-tpp.tex
\documentclass{article}

\usepackage{authblk}
\usepackage{fullpage}

\usepackage[numbers,sort&compress,square]{natbib}
\usepackage{stackengine}

\usepackage[ruled,linesnumbered]{algorithm2e}
\usepackage{bbm}
\usepackage{amsmath,amssymb}
\usepackage{enumitem}
\usepackage{bm}
\usepackage{Definitions}
\usepackage{graphicx}
\usepackage{tikz}
\usepackage{subcaption}
\usepackage{caption}
\usepackage{xcolor}
\newcommand{\xhdr}[1]{\vspace{1mm} \noindent{{\bf #1.}}}

\title{Counterfactual Temporal Point Processes}

\author[1]{Kimia Noorbakhsh\thanks{This work was done during Kimia Noorbakhsh'{}s internship at MPI-SWS.}}
\author[2]{Manuel Gomez Rodriguez}

\affil[1]{Sharif University of Technology, kimianoorbakhsh@gmail.com}
\affil[2]{Max Planck Institute for Software Systems, manuelgr@mpi-sws.org}

\date{}

\begin{document}

\maketitle

\begin{abstract}
\input{000abstract}
\end{abstract}

\section{Introduction}
\label{sec:introduction}
\input{010introduction}

\section{Preliminaries}
\label{sec:preliminaries}
\input{020preliminaries}

\section{A Causal Model of Thinning}
\label{sec:thinning}
\input{030thinning}

\section{Sampling Counterfactual Events}
\label{sec:superposition}
\input{040superposition}

\section{Experiments on Synthetic Data}
\label{sec:synthetic}
\input{050synthetic}

\section{Experiments on Real Data}
\label{sec:real}
\input{060real}

\section{Conclusions}
\label{sec:conclusions}
\input{070conclusions}

\section{Acknowledgements}
\label{sec:acknowledgements}
We would like to thank William Trouleau for sharing with us a pre-processed version of the Ebola dataset as well as the fitted stochastic block
model we used in our simulation experiments. 
Gomez-Rodriguez acknowledges support from the European Research Council (ERC) under the European Union'{}s Horizon 2020 research and
innovation programme (grant agreement No. 945719).

\bibliographystyle{unsrt}
\bibliography{refs}

\clearpage
\newpage

\appendix
\input{080appendix.tex}

\end{document}

%% file: 000abstract.tex
Machine learning models based on temporal point processes are the state of the art in a wide 
variety of applications involving discrete events in continuous time.
However, these models lack the ability to answer counterfactual questions, which are 
increasingly relevant as these models are being used to inform targeted interventions.
In this work, our goal is to fill this gap.
To this end, we first develop a causal model of thinning for temporal point processes that builds 
upon the Gumbel-Max structu\-ral causal model.
This model satisfies a desira\-ble counterfactual monotonicity condition, which is sufficient to identify 
counterfactual dynamics in the process of thinning.
Then, given an observed realization of a temporal point process with a given intensity function, 
we develop a sampling algorithm that uses the above causal model of thinning and the superposition theorem
to simulate counterfactual realizations of the temporal point process under a given alternative intensity 
function.
Simulation expe\-ri\-ments using synthetic and real epi\-de\-mio\-lo\-gi\-cal data show that the counterfactual 
rea\-li\-za\-tions provided by our algorithm may give valuable insights to enhance targeted interventions.

%% file: 010introduction.tex
In recent years, machine learning models based on temporal point processes have become increasingly popular for modeling discrete event 
data in con\-ti\-nuous time~\citep{rodriguez2018learning, yan2019modeling}. 
This type of data is ubiquitous in a wide range of application domains, from social and information networks to finance or epidemiology.  
For example, in social and information networks, events may represent users'{} posts, clicks or likes~\citep{farajtabar2017coevolve}; 
in finance, they may represent buying and selling orders~\citep{linderman2014discovering}; 
or, in epidemiology, they may represent when an individual gets infected or recovers~\citep{kim2019modeling}. 
In many of these domains, these models have become state of the art at predicting future events given a sequence of past events~\citep{shchur2021neural}.

Building upon the above models, a recent line of work~\citep{zarezade2017steering, upadhyay2018deep, tabibian2019enhancing} has developed
machine lear\-ning methods to automate online, adaptive targeted interventions using rein\-for\-cement lear\-ning and stochastic optimal control.
While this line of work has shown ear\-ly promise, particularly in personalized teaching and viral marketing, there are many high-stakes application 
in which targeted interventions are unlikely to be automated. 
For example, in epidemiology, fine-grained interventions that are targeted at particular sites or individuals (\eg, hygienic measures at work places, closures 
of schools, or contact tracing) are likely to be decided by governments, policy makers and health authorities, at least in the foreseeable future.
In this work, our goal is to develop machine learning me\-thods that are able to assist decision makers at im\-ple\-men\-ting interventions in these 
high-stakes applications.

More specifically, we focus on facilitating counterfactual thinking, a type of thinking that has been argued to play a key role in human decision 
making~\citep{byrne2002mental, epstude2008functional}.
In counterfactual thin\-king, given a history of past events that have already occurred, one asks what past events would have instead occurred if
certain intervention had been in place. 
For example, in epidemiology, assume that, du\-ring a pandemic, a government decides to implement business restrictions every time the 
weekly incidence---the (relative) number of new cases---is larger than certain threshold but unfortunately the incidence nevertheless spirals 
out of control. 
In this case, counterfactual thinking would help the government understand retrospectively to what extent the incidence would have grown had a lower threshold been implemented\footnote{Note that existing epidemiological models~\citep{hethcote2000mathematics}, in\-clu\-ding those developed in the context of COVID-19~\citep{bertozzi2020challenges, chang2021mobility, kucharski2020effectiveness, lorch2020quantifying, wells2020impact}, are unable to answer such counterfactual questions---they can only predict what the (average) future would look like under certain interventions given the past.}.

\xhdr{Our contributions} Our starting point is Lewis' thi\-nning algorithm~\citep{lewis1979simulation}, one of the most popular techniques for sampling 
in the temporal point process literature. 
Lewis' thinning algorithm first samples a sequence of potential events from a temporal point process with a constant intensity that upper 
bounds the intensity of the temporal point process of interest.
Then, it accepts each of these events with pro\-ba\-bi\-li\-ty proportional to the ratio between the intensity of the temporal point process
of interest at the time of the sampled event and the constant intensity.
%
%
In our work, the key idea is to augment the above thi\-nning process using a particular class of structural equation models (SCMs)---the Gumbel-Max 
structural causal model~\citep{oberst2019counterfactual}.
This causal model satisfies a desirable mo\-no\-to\-ni\-ci\-ty condition and, given a sequence of accepted and rejected events, it allows\- us to reliably 
estimate what events would have been accepted and rejected under an alternative intensity.
Put differently, it can be used to answer counterfactual questions about a set of previously accepted and rejected events by the thinning algorithm.

Unfortunately, the above causal model on its own is not sufficient to answer counterfactual questions about observed sequence of real events.
This is because, in ge\-ne\-ral, real event data is not \emph{generated} by a thinning process and, as a consequence, it does not \emph{include} 
rejected events.
However,  we are able to overcome this limitation by using the superposition theorem~\citep{kingman1992poisson} to sample \emph{plausible} sequences of events that the thinning process would have rejected if it had accepted the observed sequence of real events. 
Then, these generated sequences of rejected events, together with the observed events, can be fed into the above causal model of thinning to 
sample counterfactual events given an alternative intensity. 
Importantly, while our causal model of thinning only allows for inhomogeneous Poisson processes (\ie, it requires the intensities of 
interest to be deterministic), 
we can still use it to sample counterfactual events from linear Hawkes processes~\citep{hawkes1971spectra}, a popular type of temporal 
point processes with stochastic self-exciting intensities, by carefully exploiting their branching process interpretation~\citep{hawkes1974cluster}.

Finally, we evaluate our sampling algorithm using both synthetic and real epidemiological data and show that the counterfactual events 
provided by our algorithm can give valuable insights to enhance targeted interventions\footnote{To facilitate research in this area, we release 
an open-source implementation of our algorithms and data at \href{https://github.com/Networks-Learning/counterfactual-ttp}{https://github.com/Networks-Learning/counterfactual-ttp}.}.

\xhdr{Further related work} 
The literature on temporal point processes related to causal inference has mostly focused on measuring causal influence by means of, \eg, 
Granger cau\-sa\-li\-ty~\citep{xu2016learning, zhang2020cause, cai2021thp}, integrated cumulants~\citep{achab2017uncovering, trouleau2021cumulants} 
or Wold processes~\citep{figueiredo2018fast, etesami2021variational}, and on predicting quantities related to an interventional distribution of 
interest~\cite{gao2021causal, didelez2015causal, aalen2020time, lok2008statistical}
However, there exist a few notable exceptions, which have focused on counterfactual reasoning~\cite{schulam2017reliable, ryalen2020causal, roysland2012counterfactual}.
The work by Schulam and Saria~\cite{schulam2017reliable} focuses on reasoning about the counterfactual distributions of the marks in a marked temporal point process, 
rather than reasoning about the counterfactual intensities of the events as we do.
The works by Ryalen et al.~\cite{ryalen2020causal} and Roysland~\cite{roysland2012counterfactual} focus on survival analysis, \ie, temporal point processes
that terminate after one event, and are very different to ours at a technical level. In contrast, we focus on temporal point processes with multiple events.

More broadly, the literature on causal inference has a long and rich history~\citep{imbens2015causal}. However, most of this li\-te\-ra\-tu\-re has used counterfactual 
rea\-so\-ning to predict quantities related to the interventional distribution of interest such as, \eg, the conditional average treatment effect (CATE).
Two recent notable exceptions are by Oberst and Sontag~\cite{oberst2019counterfactual} and Tsirtsis et al.~\cite{tsirtsis2021counterfactual}, which have used the 
Gumbel-Max structural causal model to reason about counterfactual distributions in Markov decision processes (MDPs).
However, to the best of our knowledge, the Gumbel-Max structural causal model has not been used previously to reason about counterfactual 
events in temporal point processes.
%

%% file: 020preliminaries.tex
In this section, we first briefly revisit the frameworks of temporal point processes~\citep{daley2003introduction} and structural 
causal models~\citep{peters2017elements}.

\xhdr{Temporal point processes} 
A temporal point process is a stochastic process whose realization consists of a sequence of discrete events
localized in continuous time, $\Hcal = \{ t_i \in \RR^{+} \given i \in \NN^{+}, t_i < t_{i+1} \}$. 
A temporal point process can be equivalently represented as a counting process, $N(t)$, which records the number
of events before time $t$.
Moreover, in an infinitesimally small time window $dt$ around time $t$, it is assumed that only one event can happen, \ie,
$dN(t) \in \{0, 1\}$.

A temporal point process is typically characterized via its intensity function $\lambda(t) \geq 0$, which determines the 
probability of observing an event in $[t, t+dt)$, \ie,
\begin{equation}
\lambda(t) dt = \PP\{ dN(t) = 1 \} = \EE[dN(t)].
\end{equation}
%
In general, the intensity $\lambda(t)$ may depend on the history $\Hcal(t) = \{ t_i \in \Hcal \given t_i < t \}$
up to time $t$ and its functional form is often chosen to capture the phenomena of interest.
Throughout the paper, we will consider inhomogenous Poisson processes, \ie, $\lambda(t) = g(t)$,
where $g(t) \geq 0$ is a time-varying function, and linear Hawkes processes, \ie,
\begin{equation} \label{eq:hawkes}
\lambda(t) = \mu + \alpha \sum_{t_i \in \Hcal(t)} g(t - t_i),
\end{equation}
where $\mu \geq 0$ and the second term, with $\alpha \geq 0$, $g(t) \geq 0$ and $g(t) = 0$ for all $t < 0$, denotes the influence of 
pre\-vious events on the current intensity\footnote{The function $g(t)$ is often called triggering kernel.}.

\xhdr{Structural causal models}
Given a set of random variables $\Xb = \{ X_1, \ldots, X_n \}$, a structural causal model (SCM) $\Ccal$ defines a complete data-generating
process via a collection of assignments
\begin{equation}
X_i := f_i(\mathbf{PA}_i, U_i), 
\end{equation}
where $\mathbf{PA}_i \subseteq \Xb \backslash X_i$ are the direct causes of $X_i$, $\Ub = \{U_1, \ldots, U_n\}$ are jointly independent 
noise va\-ria\-bles, and $P(\Ub)$ denotes the (prior) distribution of the noise variables.
Here, note that, given an observational distribution $P(X_1, \ldots, X_n)$, there always exists a distribution for the noise va\-ria\-bles 
and functions $f_i$ so that $P = P^{\Ccal}$, where $P^{\Ccal}$ is the distribution entailed by $\Ccal$. 

Given a SCM $\Ccal$, we can express (atomic) interventions $\Ical$ using the \emph{do-operator}, \eg, $\Ical = \text{do}(X_i = x)$ 
corresponds to replacing the causal mechanism $f_i(\mathbf{PA}_i, U_i)$ with $x$. 
The intervened SCM is typically denoted as $\Ccal^{\Ical}$ and the interventional distribution entailed by the intervened SCM as 
$P^{\Ccal \,;\, \Ical}$.
Moreover, given a SCM $\Ccal$ and an observed realization of assignments $\Xb = \xb$, we can define a counterfactual SCM 
$\Ccal_{\Xb = \xb}$ where the noise $\Ub$ variables are distributed according to the posterior distribution $P(\Ub \given \Xb = \xb)$ 
and not necessarily jointly independent anymore.
Counterfactual statements can now be seen as interventions in a counterfactual SCM $\Ccal_{\Xb = \xb}$ and, given an intervention $\Ical$, 
we denote the interventional counterfactual distribution entailed by $\Ccal^{\Ical}_{\Xb = \xb}$ as $P^{\Ccal \given \Xb = \xb \,;\, \Ical}$.
However, the posterior distribution of the noise variables may be non-identifiable without further assumptions. 
This is because there may be several noise distributions and functions $g_i$ consistent with the observational distribution but result in 
different counterfactual distributions. 

In the context of binary random variables, mo\-no\-to\-ni\-ci\-ty is an assumption that avoids the above mentioned non-identifiability issues---it 
restricts the class of possible SCMs to those which all yield equivalent counterfactual distributions over a binary variable of in\-te\-rest~\citep{pearl2000models, oberst2019counterfactual}.
More specifically, a SCM $\Ccal$ of a binary variable $Y$ is monotonic with respect to a binary variable $T$ if and only if the 
condition
\begin{equation*}
P^{\Ccal \,;\, \text{do}(T = t)}(Y = y) \geq P^{\Ccal \,;\, \text{do}(T = t')}(Y = y)
\end{equation*}
implies that $P^{\Ccal \given Y = y,\, T = t' \,;\, \text{do}(T = t)}(Y = y')  = 0$, where $y' \neq y$.

%% file: 030thinning.tex
In this section, we first revisit Lewis'{} thinning algorithm~\citep{lewis1979simulation}, one of the most popular techniques to 
simulate event data from inhomogeneous Poisson processes. 
Then, we augment this cla\-ssi\-cal algorithm using a particular class of SCMs satisfying the monotonicity assumption, the 
Gumbel-Max SCMs~\citep{oberst2019counterfactual}. 
Finally, we demonstrate that the resulting algorithm can be used to answer counterfactual questions about a set of previously 
si\-mu\-la\-ted events.

Let $\Mcal$ be a set of inhomogeneous Poisson processes of interest and, for each $m \in \Mcal$, assume its corres\-pon\-ding 
intensity function $\lambda_{m}(t) \leq \lambda_{\max}$ for all $t \in \RR^{+}$.
To sample a sequence of events $\Hcal_{m}$ from any process $m \in \Mcal$, Lewis' thinning algorithm first samples a sequence 
of potential events $\Hcal_{\max}$ from a homogenous Poisson process with intensity $\lambda_{\max}$. 
Then, for each event $t_i \in \Hcal_{\max}$, it additionally samples a Bernoulli random variable $X_i$ with parameter 
$p = p(\lambda_{m}(t_i)) = \lambda_{m}(t_i) / \lambda_{\max}$. 
Finally, it accepts all the events $t_i$ such that $X_i = 1$.
%
\ie, $\Hcal_{m} = \{ t_i \in \Hcal_{\max} \given X_i = 1 \}$.
Here, note that the specific choice of $\lambda_{\max}$ does not affect the distribution of accepted events as long as $\lambda_{\max} \geq \lambda_m(t)$ 
for all $m \in \Mcal$ and $t \in \RR^{+}$.\footnote{In this context, note that, rather than using an homogeneous Poisson process with intensity $\lambda_{\max}$, 
one could use any process with (time-varying) intensity $\lambda'(t) \geq \lambda_m(t)$ for all $m \in \Mcal$ and $t \in \RR^{+}$, as shown in Theorem~1 in Lewis and 
Shedler~\cite{lewis1979simulation}.}
Algorithm~\ref{alg:lewis} in Appendix~\ref{app:lewis} summarizes the overall procedure, where the parameter $p$ is often 
called the thinning probability.

Given a process of interest with intensity $\lambda_m(t)$, Lewis' thinning algorithm is helpful to make predictions about future events.
For example, it can be used to compute Monte Carlo estimates of the average number of events $\EE[N(t)]$ at a time $t$ in the future. 
However, it is not sufficient to make counterfactual predictions, \eg, given the sequences of events $\Hcal_{\max}$ and $\Hcal_{m}$, we cannot know 
what would have happened if, at time $t_i$, the intensity had been $\lambda_{m'}(t_i)$, with $m \neq m'$, instead of $\lambda_m(t_i)$.
To overcome this limitation, we will now augment the above thinning algorithm using a Gumbel-Max SCM.
More specifically, let $\Ccal$ be a SCM defined by the assignments:
\begin{equation}
X_i = \argmax_{x \in \{0, 1\}} \, g(x, \Lambda_i, \Ub_i) \quad \text{and} \quad \Lambda_i = \lambda(t_i),
\label{eq:gumbel-max}
\end{equation}
where 
\begin{equation*}
g(x, \Lambda_i, \Ub_i) = \log p(X_i = x \given \Lambda_i) + U_{i, x},
\end{equation*}
with $p(X_i = x \given \Lambda_i) = x \, p(\Lambda_i) + (1-x) \, (1 - p(\Lambda_i))$, $p(\Lambda_i) = \Lambda_i / \lambda_{\max}$, 
$U_{i, x} \sim \mathrm{Gumbel}(0,1)$, 
and $t_i \sim \lambda_{max}$.
Then, the thinning probabilities in Lewis' thinning algorithm are given by the following interventional distributions over $\Ccal$\footnote{This equality follows immediately from 
the Gumbel-Max \emph{trick}~\citep{luce1959individual}}:
\begin{equation} \label{eq:distribution}
P^{\Ccal \,;\, \text{do}(\Lambda_i = \lambda(t_i))}(X_i = 1) = p(\lambda(t_i)) = \frac{\lambda(t_i)}{\lambda_{\max}}.
\end{equation}

Under this view, given a sequence of accepted events $\Hcal_m$ and rejected events $\Hcal_{\max} \backslash \Hcal_m$ under an 
inten\-si\-ty $\lambda_{m}(t)$, as determined by the binary samples $\{ x_i \}$,
we can estimate the posterior distribution $P^{\Ccal \given X_i = x_i, \, \Lambda_i = \lambda_{m}(t_i) \,;\, \text{do}(\Lambda_i = \lambda_{m'}(t_i))}(U_{i, x})$ 
of each Gumbel noise variable $U_{i, x}$ using an efficient procedure, described elsewhere~\citep{gumbelmachinery, oberst2019counterfactual}. 
Then, we can use these noise posterior distributions to compute an un\-biased finite sample Monte-Carlo estimate of the counterfactual thinning probability, 
\ie,
\begin{equation*}
P^{\Ccal \given X_i = x_i, \Lambda_i = \lambda_{m}(t_i) \,;\, \text{do}(\Lambda_i = \lambda_{m'}(t_i))}(X_i = x) 
= \EE_{\Ub_{i} \given X_i = x_i, \Lambda_i = \lambda_m(t_i)}[\mathbf{1}[x = \argmax_{x' \in \{0, 1\}} \, g(x', \lambda_{m'}(t_i), \Ub_i)]], 
\end{equation*}
where we drop $\text{do}(\cdot)$ because $\Ub_{i}$ and $\lambda_{m}(t_i)$ are independent in the counterfactual SCM.
Importantly, the above counterfactual thinning pro\-ba\-bi\-li\-ty allows us to make counterfactual predictions, \eg, given a sequence of accepted events $\Hcal_m$ 
and rejected events $\Hcal \backslash \Hcal_m$ under intensity $\lambda_m(t)$, we can use the counterfactual thinning probability to predict which events, among
those in $\Hcal$, would have been accepted if the intensity had been $\lambda_{m'}(t)$ instead of $\lambda_m(t)$.
Algorithm~\ref{alg:counterfactual-lewis} summarizes the resulting algorithm.
Here, it is important to note that the specific choice of $\lambda_{\max}$ does not affect the distribution of counterfactual events as long as $\lambda_{\max} \geq \lambda_m(t)$ 
for all $m \in \Mcal$ and $t \in \RR^{+}$, similarly as in standard thinning (refer to Appendix~\ref{app:invariance} for a proof).
\IncMargin{1.2em}
\begin{algorithm}[t]
\small
\SetKwProg{Fn}{function}{:}{end}
\textbf{Input}: $\lambda_m(t)$, $\lambda_{m'}(t)$, $\Hcal_m$, $\Hcal_{\max}$, $\lambda_{\max}$. \\
\textbf{Initialize}: $\Hcal_{m'} = \emptyset$. \\

\vspace{2mm}
\Fn{$\textsc{Acc}(\lambda_m(t), \lambda_{m'}(t), \Hcal_m, \Hcal_{\max}, \lambda_{\max})$} {
$\Hcal_{m'} \leftarrow \emptyset$ \\
\For{$t_i \in \Hcal_{\max}$}{
  $x_i \leftarrow \mathbf{1}[t_i \in \Hcal_m]$ \\
  $x'_i \sim P^{\Ccal \given X_i = x_i, \Lambda_i = \lambda_{m}(t_i) \,;\, \text{do}(\Lambda_i = \lambda_{m'}(t_i))}(X)$ \\ 
  \If{$x'_i = 1$} {
  	$\Hcal_{m'} \leftarrow \Hcal_{m'} \cup \{ t_i \}$ \\
  }
}
\textbf{Return} $\Hcal_{m'}$
}
\caption{It samples a counterfactual sequence of accepted events given a sequence of accepted and rejected events
provided by Lewis' thinning algorithm} \label{alg:counterfactual-lewis}
\end{algorithm}
\DecMargin{1.2em}

Finally, it is important to note that, by using Gumbel-Max SCMs, our causal model of thinning satisfies the monotonicity assumption~\citep{pearl2000models, oberst2019counterfactual},
discussed in Section~\ref{sec:preliminaries}, and thus the counterfactual thinning probability does not suffer from non-identifiability issues. More formally, we have the follo\-wing 
proposition (proven in Appendix~\ref{app:prop:monotonicity}):
\begin{proposition} \label{prop:monotonicity}
Let $\Mcal = \{ \lambda_m(t), \lambda_{m'}(t) \}$ and $\Ccal$ be the corresponding causal model of thinning, as defined in Eq.~\ref{eq:gumbel-max}.
Then, if $\lambda_m(t_i) \geq \lambda_{m'}(t_i)$, it holds that 
\begin{equation*}
P^{\Ccal \given X_i = 0, \Lambda_i = \lambda_{m}(t_i) \,;\, \text{do}(\Lambda_i = \lambda_{m'}(t_i))}(X_i = 1) = 0.
\end{equation*}
Conversely, if $\lambda_m(t_i) \leq \lambda_{m'}(t_i)$, it holds that 
\begin{equation*}
P^{\Ccal \given X_i = 1, \Lambda_i = \lambda_{m}(t_i) \,;\, \text{do}(\Lambda_i = \lambda_{m'}(t_i))}(X_i = 0) = 0.
\end{equation*}
\end{proposition}
The above result directly implies that, if a potential event $t_i \in \Hcal_{\max}$ was rejected under $\lambda_{m}(t)$, \ie, $t_i \notin \Hcal_m$, then in a counterfactual 
scenario, it is \emph{unlikely} that the event is accepted under $\lambda_{m'}(t)$, \ie, $t_i \in \Hcal_{m'}$, if $\lambda_{m'}(t) \leq \lambda_{m}(t)$. 
Conversely, if a potential event $t_i \in \Hcal_{\max}$ was accepted under $\lambda_{m}(t)$, \ie, $t_i \in \Hcal_m$, then in a counterfactual scenario, it is \emph{unlikely}
that the event is rejected under $\lambda_{m'}(t)$, \ie, $t_i \notin \Hcal_{m'}$, if $\lambda_{m'}(t) \geq \lambda_{m}(t)$. 

%% file: 040superposition.tex
In this section, we develop a sampling algorithm that, given an observed realization from a temporal point process with 
a given intensity function, it uses Algorithm~\ref{alg:counterfactual-lewis} and the superposition theorem~\citep{kingman1992poisson} 
to generate counterfactual realizations of the temporal point process under a given alternative intensity function. 
To ease the exposition, we first focus on inhomogeneous Poisson processes and then generalize our algorithm to linear 
Hawkes processes.

\xhdr{Inhomogenous Poisson processes} Assume we have observed a sequence of real events $\Hcal_m$ and this sequence
can be accurately characterized, observationally, using an inhomogeneous Poisson process with intensity $\lambda_{m}(t) = g(t)$,
where $g(t) \geq 0$ is a time-varying function. 
In reality, this sequence of events has not been \emph{generated} by Lewis'{} thinning algorithm but by a natural phenomena
of interest. As a result, we cannot directly apply Algorithm~\ref{alg:counterfactual-lewis} since it requires both a sequence of 
accepted and rejected events.
However, we can find plausible sequences of events that Lewis'{} thinning algorithm would have rejected if it had 
accepted the observed sequence of events.

By construction, the intensity of accep\-ted and rejected events $\lambda_{\text{accepted}}(t)$ and 
$\lambda_{\text{rejected}}(t)$ provided by Lewis'{} thinning algorithm should satisfy that
\begin{equation*}
\lambda_{\max} = \lambda_{\text{accepted}}(t) + \lambda_{\text{rejected}}(t).
\end{equation*}
Then, if $\lambda_{\text{accepted}}(t) = \lambda_m(t)$, $\lambda_{\max} \geq \max_t \, \lambda_m(t)$ and $\Hcal_{\text{accepted}} = \Hcal_m$, 
by the super\-position theorem, we can find plausible sequences of events that Lewis'{} algorithm would have rejected just by sampling 
from the intensity $\lambda_{\text{rejected}}(t) = \lambda_{\max} - \lambda_m(t)$.
Then, these generated sequences of rejected events, together with the sequence of observed events, can be fed into Algorithm~\ref{alg:counterfactual-lewis} to sample sequences of counterfactual events given an alternative intensity 
$\lambda_{m'}(t)$.
Algorithm~\ref{alg:counterfactual-inhomogeneous-poisson} summarizes the resulting algorithm, where $\textsc{Lewis}(\cdot)$ 
samples a sequence of events using Algorithm~\ref{alg:lewis} in Appendix~\ref{app:lewis} (Lewis' thinning algorithm) and
$\textsc{Acc}(\cdot)$ samples a counterfactual sequence of accepted events using Algorithm~\ref{alg:counterfactual-lewis}.
Here, note that the specific choice of $\lambda_{\max}$ does not affect the distribution of counterfactual events as long as $\lambda_{\max} \geq \lambda_m(t)$ for 
all $m \in \Mcal$ and $t \in \RR^{+}$.
\IncMargin{1.2em}
\begin{algorithm}[t]
\small
\SetKwProg{Fn}{function}{:}{end}
\textbf{Input}: $\lambda_m(t)$, $\lambda_{m'}(t)$, $\Hcal_m$, $\lambda_{\max}$, $T$. \\
\textbf{Initialize}: $\Hcal_{m'} = \emptyset$. \\

\vspace{2mm}
\Fn{$\textsc{Cf}(\lambda_m(t), \lambda_{m'}(t), \Hcal_m, \lambda_{\max}, T)$} {
$\Hcal_{\max}, \underline{\enskip} \leftarrow \textsc{Lewis}(\lambda_{\max}-\lambda_m(t), \lambda_{\max}, T)$ \\
$\Hcal_{\max} \leftarrow \Hcal_{\max} \cup \Hcal_m$ \\

$\Hcal_{m'} \leftarrow \textsc{Acc}(\lambda_m(t), \lambda_{m'}(t), \Hcal_m, \Hcal_{\max}, \lambda_{\max})$

\vspace{2mm}
\textbf{Return} $\Hcal_{m'}$
}
\caption{It samples a counterfactual sequence of events given a sequence of observed events from an inhomogeneous 
Poisson process.} \label{alg:counterfactual-inhomogeneous-poisson}
\end{algorithm}
\DecMargin{1.2em}

\xhdr{Linear Hawkes processes} Assume we have observed a sequence of real events $\Hcal_m$ and it can 
be accurately characterized, observationally, using a linear Hawkes process with an intensity $\lambda_{m}(t)$ 
pa\-ra\-me\-te\-rized by $\mu_m$, $\alpha_m$ and $g_m(\cdot)$, as defined in Eq.~\ref{eq:hawkes}. 
If our goal is to sample sequences of counterfactual events $\Hcal_{m'}$ given an alternative Hawkes intensity $\lambda_{m'}(t)$
pa\-ra\-me\-te\-rized by $\mu_{m'}$, $\alpha_{m'}$ and $g_{m'}(\cdot)$, we cannot proceed similarly as in the case of inhomogeneous 
Poisson processes.
This is because, to sample plausible rejected events within Algorithm~\ref{alg:counterfactual-inhomogeneous-poisson}, we need to pick 
a value of $\lambda_{\max}$ that upper bounds both $\lambda_{m}(t)$ and $\lambda_{m'}(t)$, otherwise, Algorithm~\ref{alg:counterfactual-lewis} might break because the thinning probabilities $p$ used by the causal model of thinning could be greater than $1$.
Unfortunately, since $\lambda_{m'}(t)$ depends on the counterfactual history $\Hcal_{m'}$ we aim to sample, we cannot know
its maximum value at the time we sample the rejected events.
However, we can overcome this challenge by resorting to the branching process interpretation of Hawkes processes~\citep{hawkes1974cluster}. 
\IncMargin{1.2em}
\begin{algorithm}[t]
\small
\textbf{Input}: $\mu_{m}$, $\alpha_{m}$, $g_{m}(t)$, $\mu_{m'}$, $\alpha_{m'}$, $g_{m'}(t)$, $\Hcal_m$, $\lambda_{\max}$, $T$. \\
\textbf{Initialize}: $\Hcal_{m'} = \emptyset$. \\

\vspace{2mm}
$\{ \Hcal_{m, j} \} \leftarrow \textsc{Assign}(\Hcal_m, \lambda_{m}(t))$
\vspace{2mm}

$\Hcal_{m', 0} \leftarrow \textsc{Cf}(\gamma_{m,0}(t), \gamma_{m',0}(t), \Hcal_{m, 0}, \lambda_{\max}, T)$ \\
$\Hcal_{m'} \leftarrow \Hcal_{m'} \cup \Hcal_{m', 0}$

\vspace{2mm}
\For{$t_j \in \Hcal_m$} {
	\If{$t_j \in \Hcal_{m'}$} {
		$\Hcal_{m', j}~\leftarrow~\textsc{Cf}(\gamma_{m,j}(t),~\gamma_{m',j}(t),~\Hcal_{m, j},~\lambda_{\max},~T)$ \\
		$\Hcal_{m'} \leftarrow \Hcal_{m'} \cup \Hcal_{m', j}$
	}
}

\vspace{2mm}
$\Hcal \leftarrow \Hcal_{m'} \backslash \Hcal_{m}$ \\
\While{$| \Hcal | > 0$} {
		$t_k \leftarrow \min_{t \in \Hcal} t$ \\
		$\Hcal_{m, k}, \underline{\enskip} \leftarrow \textsc{Lewis}(\gamma_{m', k}(t), \lambda_{\max}, T)$ \\
		$\Hcal \leftarrow \Hcal_{m, k} \cup \Hcal \backslash \{ t_k \}$ \\
		$\Hcal_{m'} \leftarrow \Hcal_{m'} \cup \Hcal_{m, k}$ \\
}

\vspace{2mm}
\textbf{Return} $\Hcal_{m'}$
\caption{It samples a counterfactual sequence of events given a sequence of observed events from a Hawkes process.} \label{alg:counterfactual-hawkes}
\end{algorithm}
\DecMargin{1.2em}

More specifically, we can view any linear Hawkes process as a superposition of several temporal point processes, \ie, 
a process with constant intensity $\gamma_{0}(t) = \mu$ and, for each event $t_i \in \Hcal$, a process with intensity 
$\gamma_{i}(t) = \alpha g(t - t_i)$.
Under this view, we can naturally derive the following thinning algorithm to sample from linear Hawkes processes~\citep{moore2018maximum}.
First, we sample a sequence of events $t_i$ from the process with intensity $\gamma_{0}(t)$. 
Then, for each sampled event $t_i$, we create a process with intensity $\gamma_{i}(t)$ and sample a sequence of events $t_j$ 
independently from each of them using Algorithm~\ref{alg:lewis} in Appendix~\ref{app:lewis} (Lewis' thinning algorithm). 
Further, for each of these sampled events $t_j$, we again create another set of processes with intensity $\gamma_{j}(t)$ and 
sample from them independently, and continue recursively.
By the superposition theorem, it readily follows that the overall sequence of sampled events is a valid rea\-li\-za\-tion of the original 
Hawkes process.
Algorithm~\ref{alg:hawkes-superposition} in Appendix~\ref{app:hawkes-superposition} summarizes the resulting algorithm.
\begin{figure*}[t]
	\begin{subfigure}[b]{0.25\textwidth}
		\centering
		\includegraphics[width=.85\linewidth]{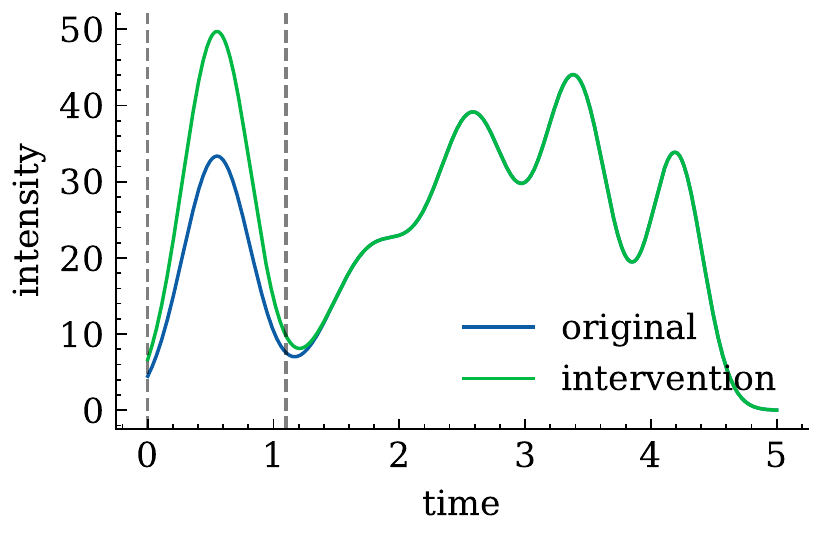}
	\end{subfigure}%
	\begin{subfigure}[b]{0.25\textwidth}
		\centering
		\includegraphics[width=.85\linewidth]{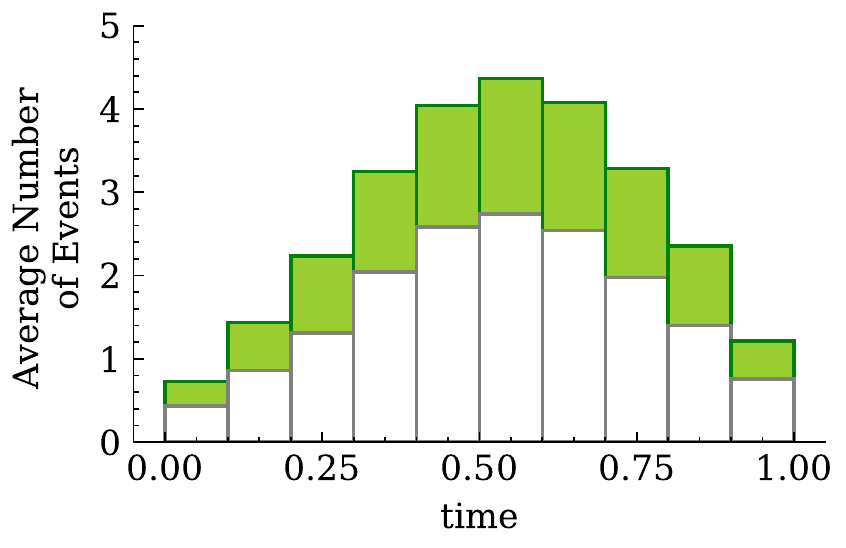}

	\end{subfigure}%
	\begin{subfigure}[b]{0.25\textwidth}
		\centering
		\includegraphics[width=.85\linewidth]{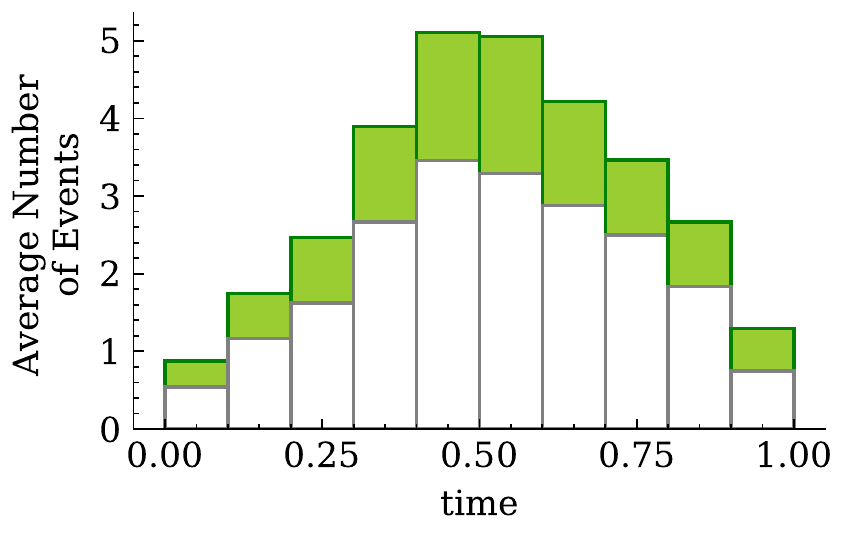}

	\end{subfigure}%
	\begin{subfigure}[b]{0.25\textwidth}
		\centering
		\includegraphics[width=.85\linewidth]{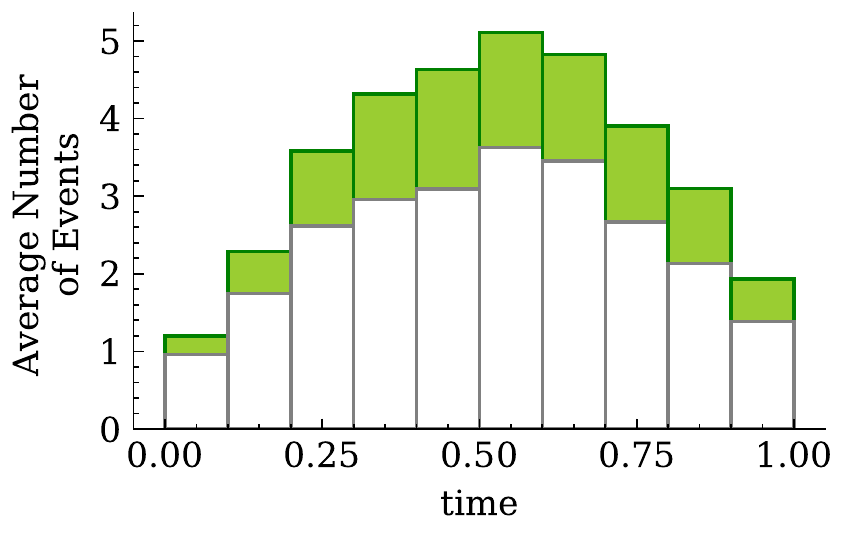}
	\end{subfigure}

	\begin{subfigure}[b]{0.25\textwidth}
		\centering
		\includegraphics[width=.85\linewidth]{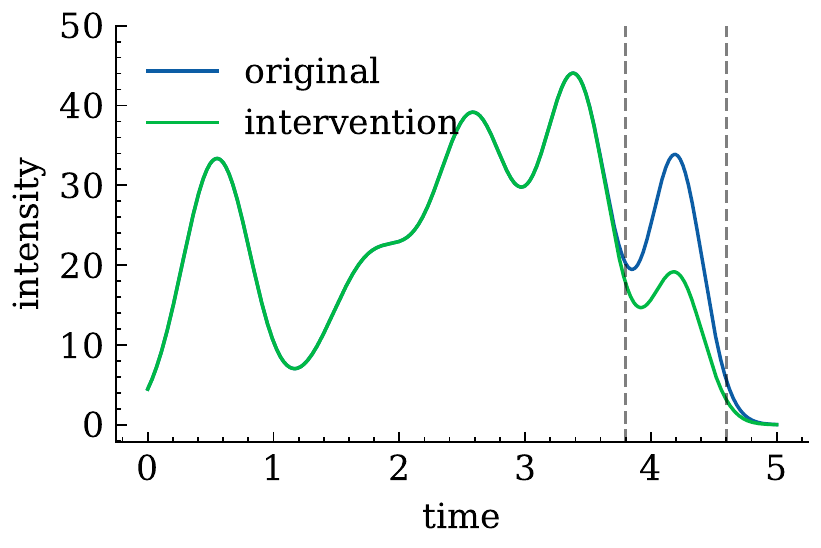}
		\caption{Intensities}
	\end{subfigure}%
	\begin{subfigure}[b]{0.25\textwidth}
		\centering
		\includegraphics[width=.85\linewidth]{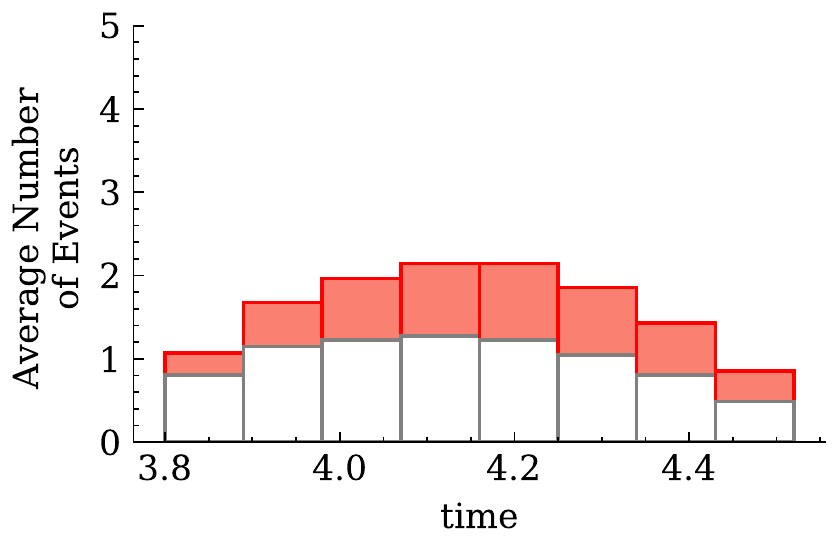}
		\caption{Low number of events}
		
	\end{subfigure}%
	\begin{subfigure}[b]{0.25\textwidth}
		\centering
		\includegraphics[width=.85\linewidth]{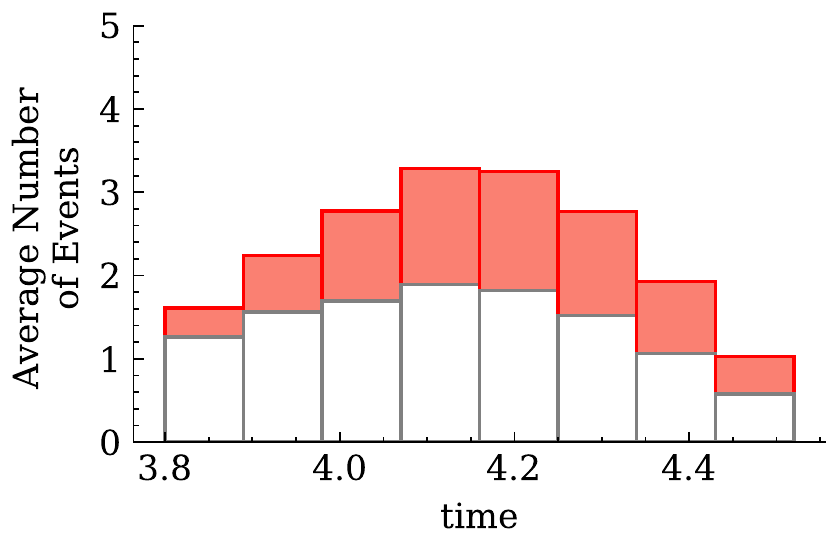}
		\caption{Medium number of events}
		
	\end{subfigure}%
	\begin{subfigure}[b]{0.25\textwidth}
		\centering
		\includegraphics[width=.85\linewidth]{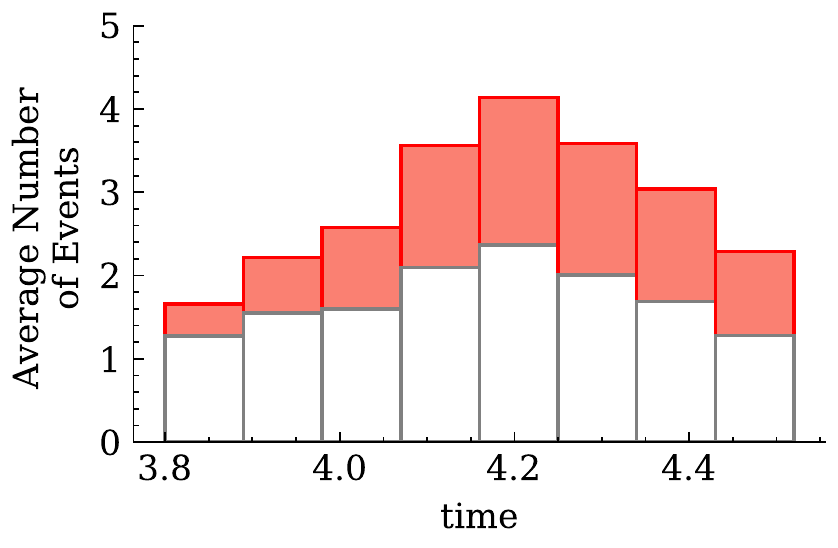}
		\caption{High number of events}
	\end{subfigure}
	\caption{Effect of interventions in two inhomogeneous Poisson processes.
	Pa\-nel (a) shows the intensities of the ori\-gi\-nal and the intervened processes (in blue and green, respectively) and a window of interest (dashed 
	vertical lines).
	Panels (b-d) show the diffe\-rence in average number of events over time between the counterfactual and the original realizations, where each row 
	corresponds to one process and we group the original realizations in three quantiles according to their overall number of events in the time window 
	of interest.
	%
	%
	Positive (negative) differences are shown in green (red).
	%
	%
	%
	In each experiment, we sample $1{,}000$ realizations from the original process and, for each of these realizations, we sample $100$ counterfactual realizations 
	from the intervened process.
	%
	}
	\label{fig:groups}
\end{figure*}

Importantly, in the above algorithm, the intensities $\gamma_{j}(t)$ of all the processes we sample from are bounded by 
$\max \{ \mu, \max_{t} \, \alpha g(t) \}$.
As a result, given two intensi\-ties of interest $\lambda_{m}(t)$ and $\lambda_{m'}(t)$, we can sample sequences of
counterfactual events by running Algorithm~\ref{alg:counterfactual-inhomogeneous-poisson} independently for each 
of the above processes with 
\begin{equation*}
\lambda_{\max} \geq \max \{ \mu_m, \mu_{m'}, \max_{t} \, \alpha_{m} g_{m}(t), \max_{t} \, \alpha_{m'} g_{m'}(t) \}.
\end{equation*}
However, to do so, we also need to assign each observed event $t_i \in \Hcal_m$ to one of the above processes with pro\-ba\-bi\-li\-ty 
$\gamma_{m,j}(t_i) / \sum_{k < i} \gamma_{m,k}(t_i)$, which is the pro\-ba\-bi\-li\-ty that the process has \emph{caused} 
the event~\citep{moore2016hawkes}.
Algorithm~\ref{alg:counterfactual-hawkes} summarizes the re\-sul\-ting algorithm, where $\textsc{Assign}(\cdot)$
returns the observed events $\Hcal_{m, j}$ assigned to each process $j$,
$\textsc{Cf}(\cdot)$ samples a counterfactual sequence of events using Algorithm~\ref{alg:counterfactual-inhomogeneous-poisson},
and $\textsc{Lewis}(\cdot)$ samples a sequence of events using Algorithm~\ref{alg:lewis} in Appendix~\ref{app:lewis} (Lewis' 
thinning algorithm).
Within the algorithm, it is also worth noting that 
lines 6-11 need to go through the observed events $t_j$ in chronological order and, for each $t_j$, the algorithm only
accepts counterfactual events from the corresponding process with intensity $\gamma_{m', j}(t)$ if the event $t_j$ has been 
previously accepted in the counterfactual realization.
Moreover, lines 12-18 recursively sample a sequence of events for each of the processes triggered by the counterfactual events 
that did not exist in the observed sequence of events.

%% file: 050synthetic.tex
In this section, we feed Algorithms~\ref{alg:counterfactual-inhomogeneous-poisson} and~\ref{alg:counterfactual-hawkes} 
with rea\-li\-za\-tions of synthetic inhomogeneous Poisson processes and linear Hawkes processes 
and investigate to what extent the counterfactual realizations returned by the algorithms under alternative 
in\-ten\-si\-ty functions differ from the ori\-gi\-nal realizations fed to them\footnote{All experiments were performed on a machine 
equipped with 48 Intel(R) Xeon(R) 3.00GHz CPU cores and 1.5TB memory.}.

\xhdr{Experimental setup} We consider the family of inho\-mogeneous Poisson processes $\Mcal(\boldsymbol{\phi}, \boldsymbol{\alpha}, \boldsymbol{\tau})$ 
pa\-ra\-me\-te\-rized by weighted combinations of RBF kernels, \ie,
\begin{equation}
\lambda(t) = \sum_j \phi_{j} \exp\left( -\alpha_{j} (t-\tau_{j}) \right), \,\, t \geq 0,
\end{equation}
where $\phi_{j}$, $\alpha_{j}$, $\tau_{j} \geq 0$, 
and the family of linear Hawkes processes $\Mcal(\mu, \alpha, \omega)$ defined in Eq.~\ref{eq:hawkes}, with exponential 
triggering kernels $g(t) = \exp(- \omega t)$.
Moreover, we experiment with simple interventions under which, for inho\-mogeneous Poisson processes, one of the RBF kernels, picked at random, 
change its amplitude, \ie, $\phi_{m', i} = \max(\phi_{m, i} + \epsilon, 0)$, and, for Hawkes processes, the parameter $\alpha$ change its 
value, \ie, $\alpha_{m'} = \max(\alpha_m + \epsilon, 0)$, where $\epsilon \sim N(0, \sigma)$.

In each experiment, we first sample $1{,}000$ realizations from a process with one set of parameters using Algorithm~\ref{alg:lewis} (or 
Algorithm~\ref{alg:hawkes-superposition}). 
Then, we carry out the above mentioned intervention and, for each of the sampled realizations, we use Algorithm~\ref{alg:counterfactual-inhomogeneous-poisson} (or Algorithm~\ref{alg:counterfactual-hawkes}) to sample $100$ counterfactual realizations under the resulting alternative set of parameters.
Finally, 
we partition the original realizations in three quantiles according to their overall number of events in a time window of interest and, for each quantile, 
we look at the number of events in the corresponding counterfactual realizations in the same window of interest. 
%
%
%
In all experiments, Algorithms~\ref{alg:counterfactual-lewis}---\ref{alg:counterfactual-hawkes} 
use $100$ samples from the posterior distribu\-tion $P^{\Ccal \given X_i=x, \Lambda_i=\lambda(t_i) \,;\, \text{do}(\Lambda_i=\lambda_{m'}(t_i))}(\Ub_{i})$ 
of each Gumbel noise variable $U_{i,x}$ to esti\-mate the counterfactual thinning pro\-ba\-bi\-li\-ties $P^{\Ccal \given X_i=x, \Lambda_i=\lambda(t_i) \,;\, \text{do}(\Lambda_i=\lambda_{m'}(t_i))}(X_i)$.
%
\begin{figure*}[!!t]	
	\begin{subfigure}{1\linewidth}
		\begin{minipage}{.35\linewidth}
		\includegraphics[width=\linewidth]{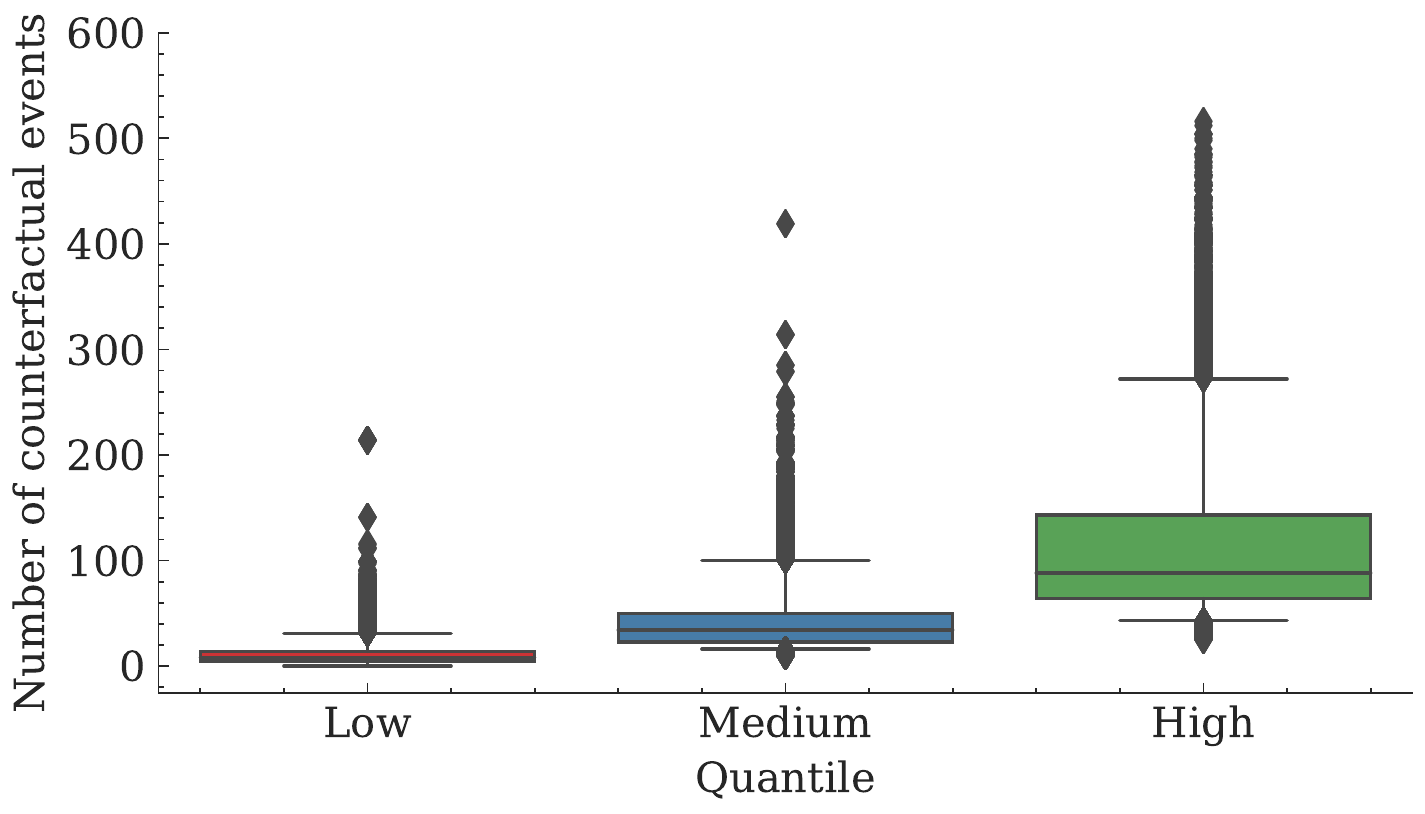}
		\centering
		\caption*{Number of counterfactual events}
		\end{minipage}%
		\hfill
		\begin{minipage}{.6\linewidth}
			\centering
			\includegraphics[width=0.85\linewidth]{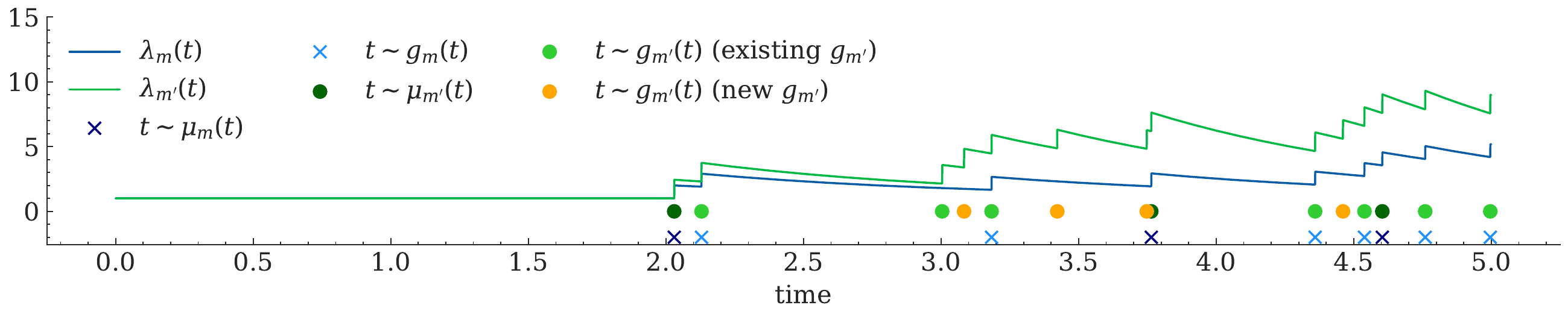}\\
			\includegraphics[width=0.85\linewidth]{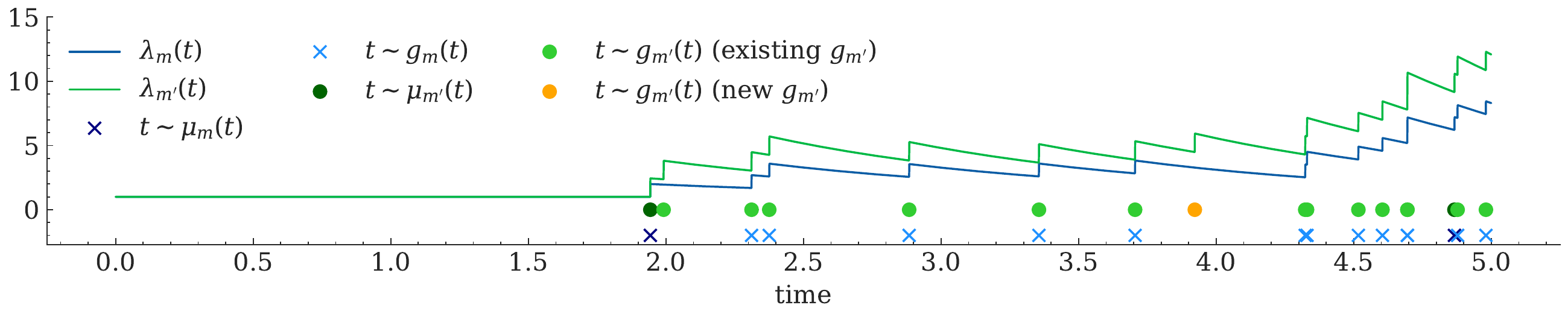}%
			\caption*{Events over time in single realizations}
		\end{minipage}
		\caption{Intervention increases self-excitation, $\alpha_{m'} > \alpha_m$}
		\vspace{5pt}
		\begin{minipage}{.35\linewidth}
			\includegraphics[width=\linewidth]{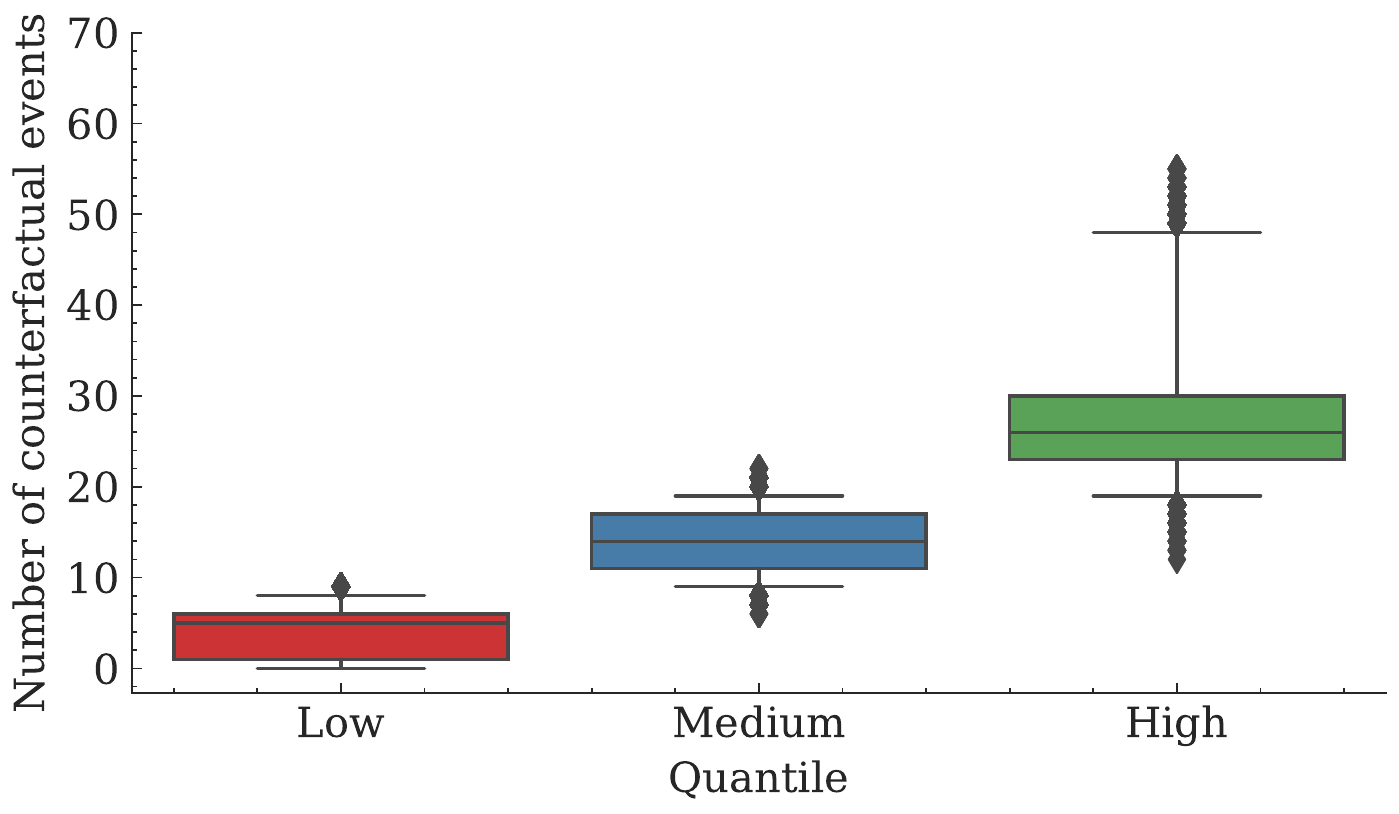}
			\centering
			\caption*{Number of counterfactual events}
		\end{minipage}%
		\hfill
		\begin{minipage}{.6\linewidth}
			\centering
			\includegraphics[width=0.85\linewidth]{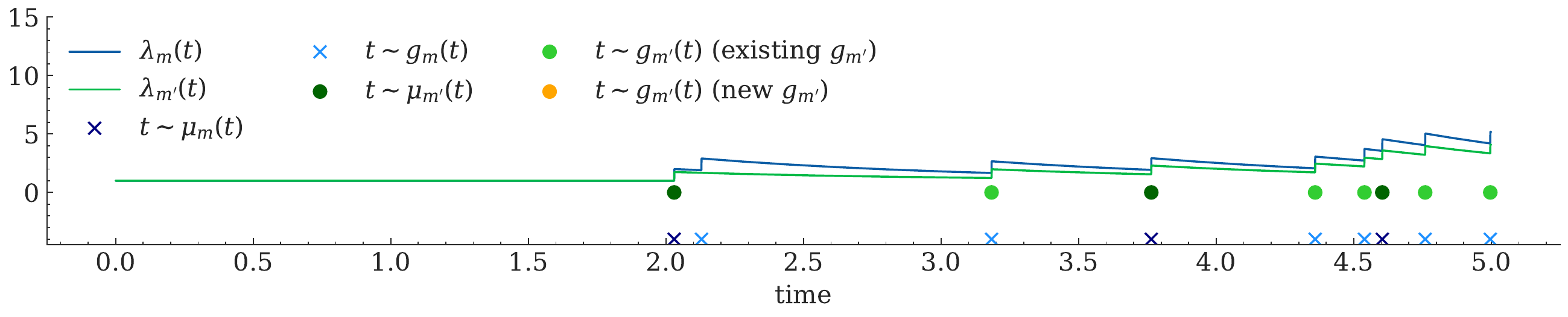}\\
			\includegraphics[width=0.85\linewidth]{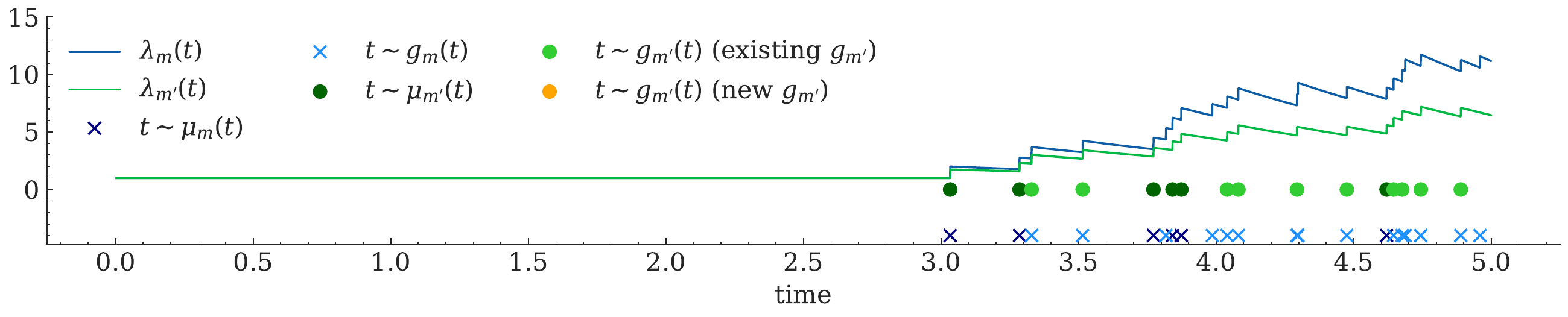}%
			\caption*{Events over time in single realizations}
		\end{minipage}
		\caption{Intervention decreases self-excitation, $\alpha_{m'} < \alpha_m$}
		\end{subfigure}
	\caption{Effect of interventions in Hawkes processes. 
	Left panels summarize the distribution of the number of events per counterfactual realization corresponding to original realizations with a low (red; $|\Hcal_m| \in [0,9]$, $\EE[|\Hcal_m|] = 5.04$), 
	medium (blue; $|\Hcal_m| \in [10,22]$, $\EE[|\Hcal_m|] = 17.17$) and high (green; $|\Hcal_m| \in [25, 59]$, $\EE[|\Hcal_m|] = 36.17$) number of events.
	The horizontal lines within the boxes indicate average value, the boxes indicate $25$\% and $75$\% quantiles, the whiskers indicate
	$5$\% and $95$\% quantiles and the points are outliers.
	Right panels show the Hawkes intensities and event times corresponding to specific realizations of the original process and the counterfactual 
	processes.
	Crosses (dots) denote events of the original (counterfactual) processes and the color pattern indicates which type of process generated each event (be it $\mu_{m'}$ or 
	$g_{m'}$ due to a counterfactual event that also existed (did not exist) in the original realization). 
	Here, we sample $1{,}000$ realizations from the original process, with parameters $\mu_{m} = 1$, $\alpha_m = 1$, and $\omega_m = 1$,
	and, for each of these realizations, we sample $100$ counterfactual realizations from each of the intervened process.
	In panel (a), the intervened process has parameter $\alpha_{m'} = 1.44$, in panel (b), it has parameter $\alpha_{m'} = 0.75$ and, in both panels, 
	the remaining parameters $\mu_{m'} = \mu_m$ and $\alpha_{m'} = \alpha_m$.
	We set the time horizon $T = 5$.
	%
	}
	\label{fig:hawkes}
\end{figure*}

\xhdr{Results} Figure~\ref{fig:groups} summarizes the results for two spe\-ci\-fic inhomogeneous Poisson processes undergoing one of the above 
mentioned interventions, which reveal several interesting insights---we found qualitatively similar results for other inhomogeneous Poisson processes.
In the top row, the results show that, under our model, an intervention that increases the original intensity of the process, by increasing the amplitude of 
one of the RFB kernels by approximately an additional half 
%
%
does not have the same effect across realizations. 
In realizations with a low (high) number of events, the average number of counterfactual events in the window of interest increases up to $\sim$$60$\% ($\sim$$40$\%)
over time with respect to the average number of events in the original realizations.
In the bottom row, we find that this is also true for an intervention that decreases the origi\-nal intensity of the process, by approximately halving the amplitude of 
another of the RBF kernels. 
However, the difference is smaller in relative terms.

Figure~\ref{fig:hawkes} summarizes the results for a specific Hawkes process undergoing two of the above mentioned interventions---we found qualitative
similar results for other Hawkes processes.
In the left panels, we find that, similarly as in inhomogeneous Poisson processes, the interventions do not have the same effect across realizations.
However, in this case, the difference among them is more stark---while the average number of coun\-ter\-fac\-tual events (a) increases by $115$\% and (b) decreases 
by $11$\% for realizations with a low number of events, it (a) increases by $216$\% and (b) decreases by $21$\% for realizations with a high number of events.
Moreover, we also find that, there is a high va\-ria\-bi\-li\-ty across counterfactual realizations, especially when $\alpha_{m'} > \alpha_m$. For example, while the original realizations 
with a low number of events never contained more than $9$ events, there exists counterfactual realizations with more than $200$ events.
This is due to the self-exciting property of Hawkes processes by which counterfactual events may trigger the emergence of additional counterfactual events, shown as yellow dots in 
the right panels. 

%

%% file: 060real.tex
In this section, we run a (simple) variation of Algorithm~\ref{alg:counterfactual-hawkes} (refer to Appendix~\ref{app:sir-counterfactuals}) to quantify
the effect of interventions on a networked Susceptible-Infectious-Recovered (SIR) epidemiological model~\cite{lorch2018stochastic} 
fitted using real event data from an Ebola outbreak in West Africa in 2013-2016~\citep{garske2017heterogeneities}.

\xhdr{Experimental setup}
We build upon the networked Susceptible-Infectious-Recovered (SIR) epidemiolo\-gi\-cal model introduced by Lorch et al.~\cite{lorch2018stochastic},
which is based on temporal point processes.
Given a contact net\-work $\Gcal = (\Vcal, \Ecal)$, we represent the times when each node gets infected and recovered using a collection of binary 
counting processes $\Yb(t)$ and $\Wb(t)$ and we track the current state of each node using a collection of state variables $\Xb(t) = \Yb(t) - \Wb(t)$,
where $X_i(t) = 1$ indicates node $i \in \Vcal$ is infected at time $t$ and $X_{i}(t) = 0$ indicates it is susceptible or recovered.
For each node $i \in \Vcal$, we characterize the above counting processes using the following (conditional) intensities
\begin{equation}
\EE[ dY_i(t) \given \Hcal(t)] = (1 - X_i(t)) \sum_{j \,|\, (i, j) \in \Ecal} \beta \, X_j(t) dt \qquad
\EE[ dW_i(t) \given \Hcal(t)] = \delta \, X_i(t) dt, \label{eq:sir}
\end{equation}
where note that the counting process $\Yb(t)$ can be viewed as a (networked) multidimensional Hawkes process with stochastic triggering 
kernels defined by step functions.
\begin{figure}[t]
	\centering
	\begin{subfigure}{.45\textwidth}
		\centering
		\includegraphics[width=0.7\linewidth]{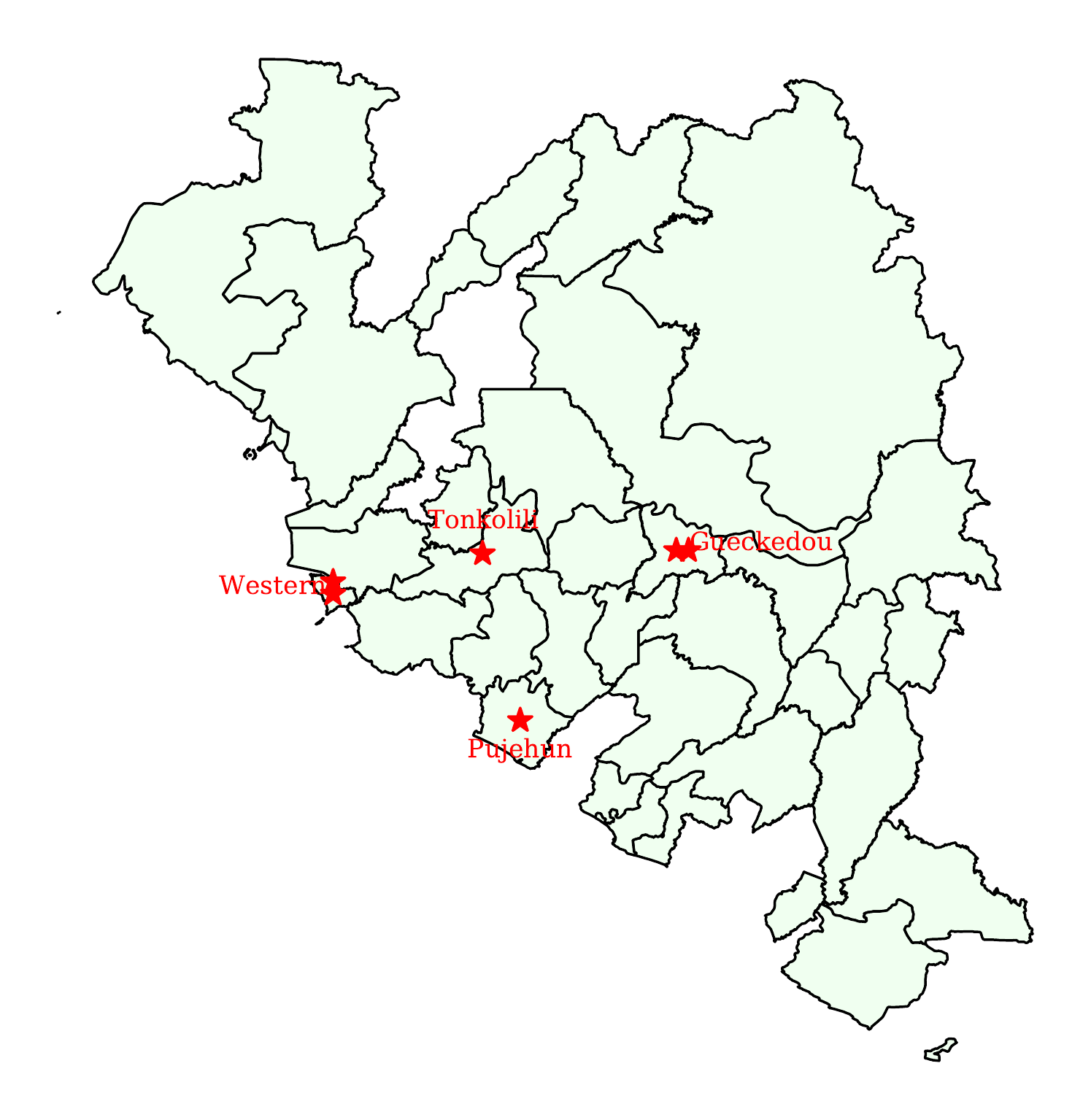}
		\caption{Seed infections}
	\end{subfigure}%
	\begin{subfigure}{.45\textwidth}
		\centering
		\includegraphics[width=0.7\linewidth]{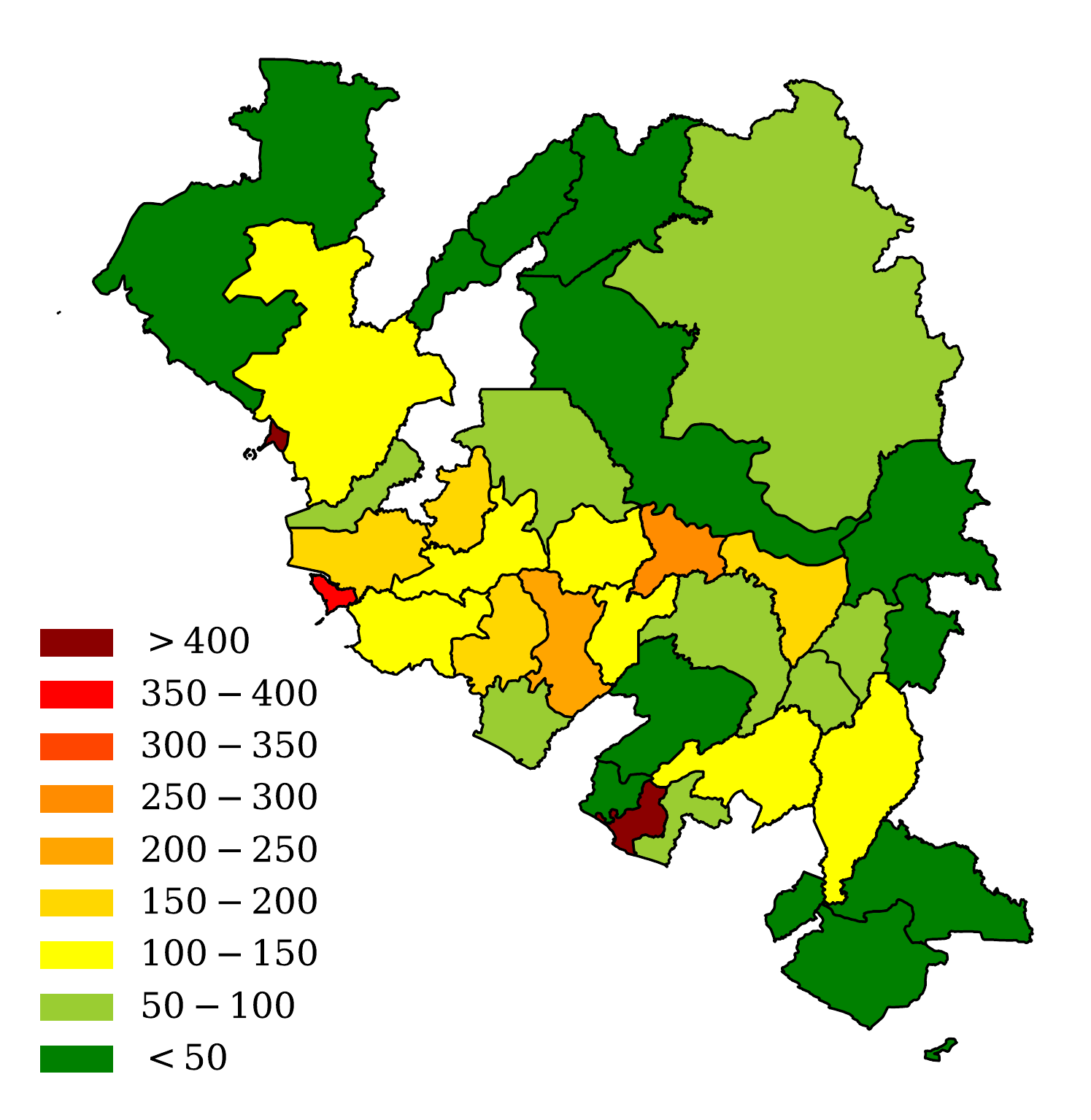}
		\caption{Overall cumulative infections}
	\end{subfigure}
	\caption{Geographical distribution of infections in a \emph{realistic} Ebola outbreak in West Africa. 
	Panel (a) shows the seed infections in each district, where each red star represents an infection.
	Panels (b) shows the overall cumulative number of infections per district.
	Darker red (green) corresponds to high (low) number of cumulative infections.
	}
	\label{fig:geo-infections}
\end{figure}

To set the model parameters $\beta$ and $\delta$, we resort to well-known epidemiological quantities whose values have been estimated by the World
Health Organization (WHO) for the Ebola outbreak in West Africa in 2013-2016~\cite{who2014ebola}. 
More specifically, the average generation time, \ie, the time between the infection of a primary case and one of its secondary cases~\cite{svensson2007note}, 
is $1/\beta \approx 15.3$ days and the mean time from the onset of symptoms to death or discharge from the hospital is $1/\delta \approx 11.4$ days.
\begin{figure}[t]
		\centering
		\begin{subfigure}{0.65\textwidth}
			\centering
			\begin{minipage}[t]{0.9\linewidth}
				\includegraphics[width=1\linewidth]{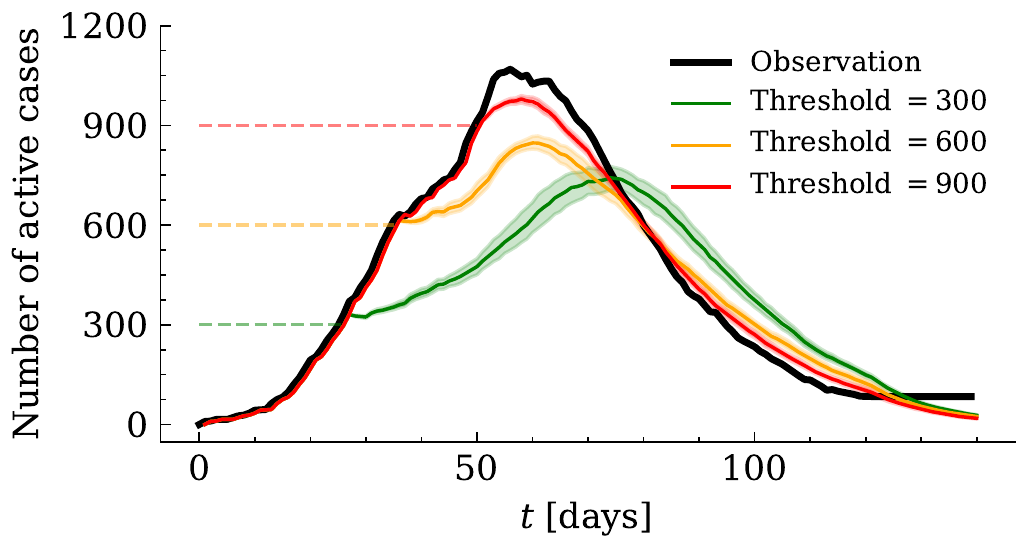}
				\caption{Reduction of contacts in the district with the highest incidence}
			\end{minipage}%
		\end{subfigure} \\ %
		\vspace{4mm}
	\begin{subfigure}{.33\textwidth}
		\centering
		\includegraphics[width=0.8\linewidth]{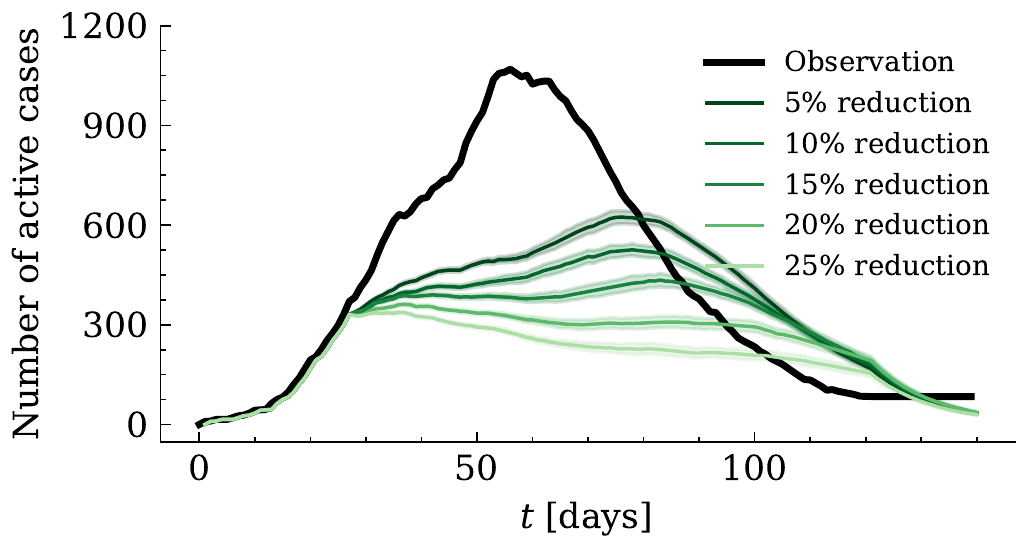}
		\caption*{Threshold = $300$}
	\end{subfigure}%
	\begin{subfigure}{.33\textwidth}
		\centering
		\includegraphics[width=0.8\linewidth]{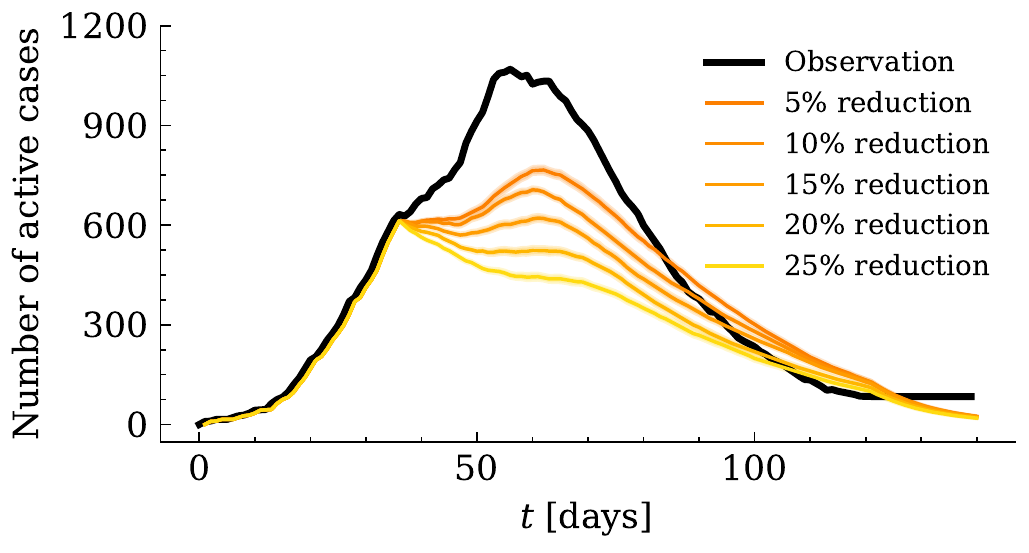}
		\caption*{Threshold = $600$}
	\end{subfigure}%
	\begin{subfigure}{.33\textwidth}
		\centering
		\includegraphics[width=0.8\linewidth]{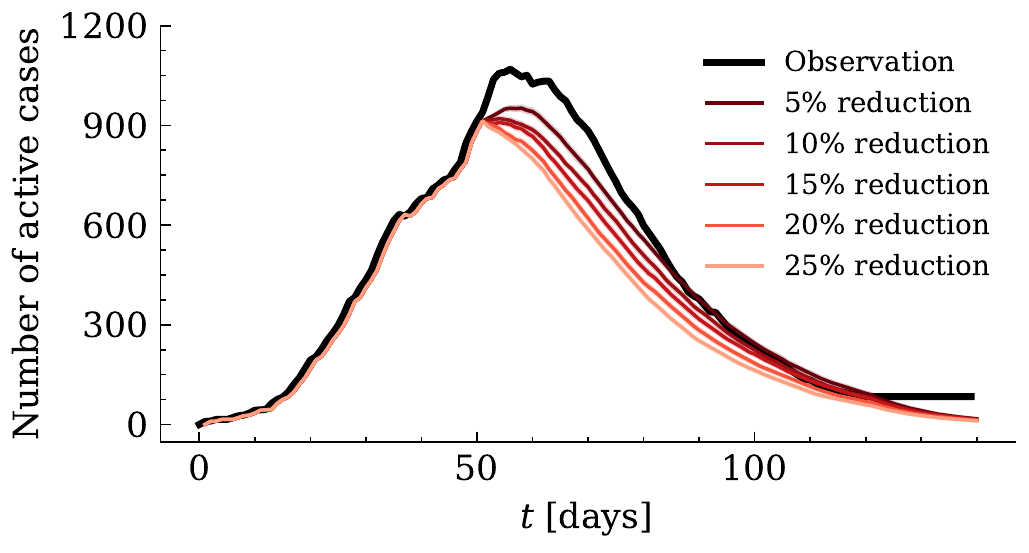}
		\caption*{Threshold = $900$}
	\end{subfigure}
\caption{Effect of interventions where individuals reduce their contacts after the overall number of active infections reaches a varying threshold 
in a \emph{realistic} Ebola outbreak in West Africa. 
In all figures, the black line (``Observation'') corresponds to an outbreak sampled from the SIR model defined by Eq.~\ref{eq:sir}, 
the remaining lines correspond to counterfactual realizations for this outbreak under different interventions,
and the shaded regions correspond to $95\%$ confidence intervals.
In panel (a), only individuals in the district with the highest incidence (at the time of the intervention) reduce their contacts by $50$\% within the district and 
get isolated from individuals from all other districts.
In panel (b), everyone reduces their individual contacts by a varying percentage.
The disease specific and network parameters of the SIR model are calibrated using data from an Ebola outbreak in West Africa in 2013-2016~\cite{garske2017heterogeneities}
and, for each threshold and reduction level, we repeat the experiment five times and, each time, sample $20$ counterfactual realizations using a variation of 
Algorithm~\ref{alg:counterfactual-hawkes} (refer to Appendix~\ref{app:sir-counterfactuals}).
%
%
}
%
\label{fig:isolation}
\end{figure}

\begin{figure}[th]
	\centering
	\begin{subfigure}{1\textwidth}
		\centering
		\includegraphics[width=0.7\linewidth]{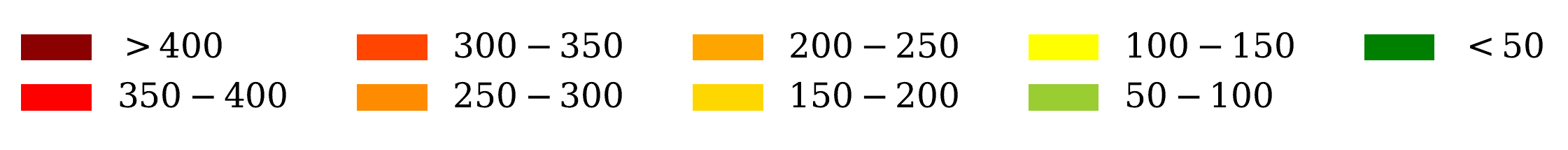}	
	\end{subfigure}
	\begin{subfigure}{.33\textwidth}
		\centering
		\includegraphics[width=0.8\linewidth]{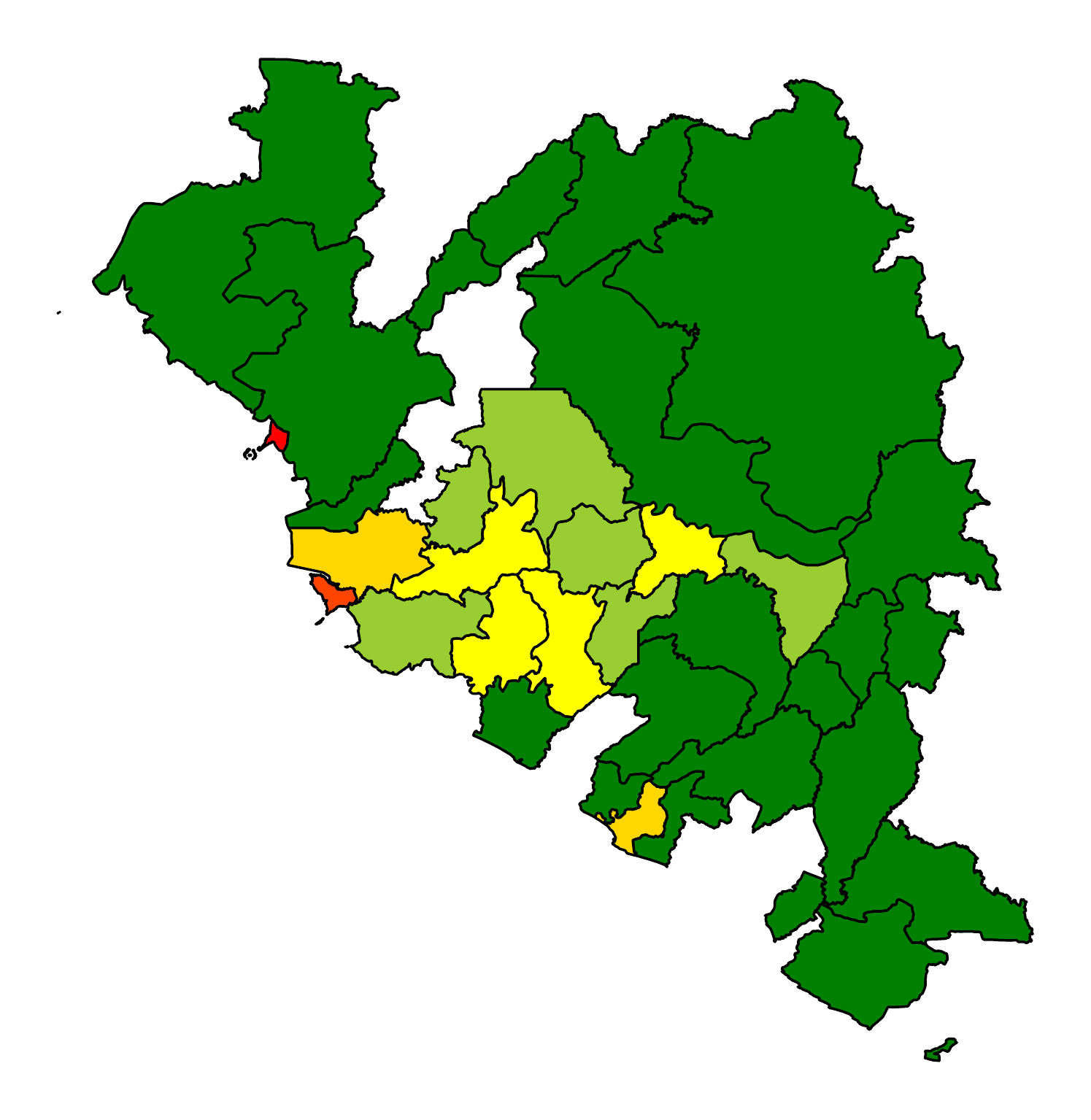}
		\caption*{Threshold = $300$}	
	\end{subfigure}%
	\begin{subfigure}{.33\textwidth}
		\centering
		\includegraphics[width=0.8\linewidth]{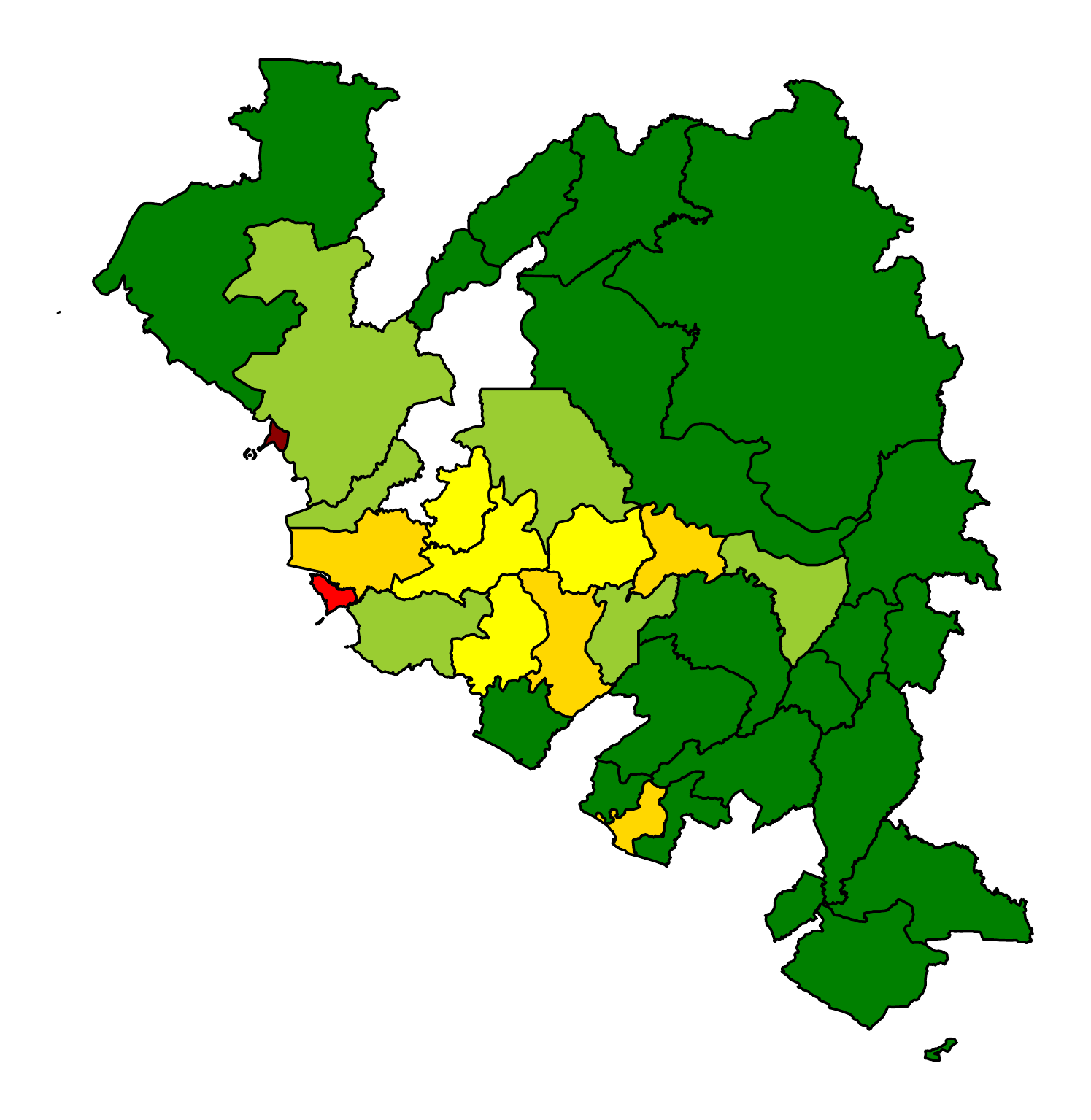}
		\caption*{Threshold = $600$}
	\end{subfigure}%
	\begin{subfigure}{.33\textwidth}
		\centering
		\includegraphics[width=0.8\linewidth]{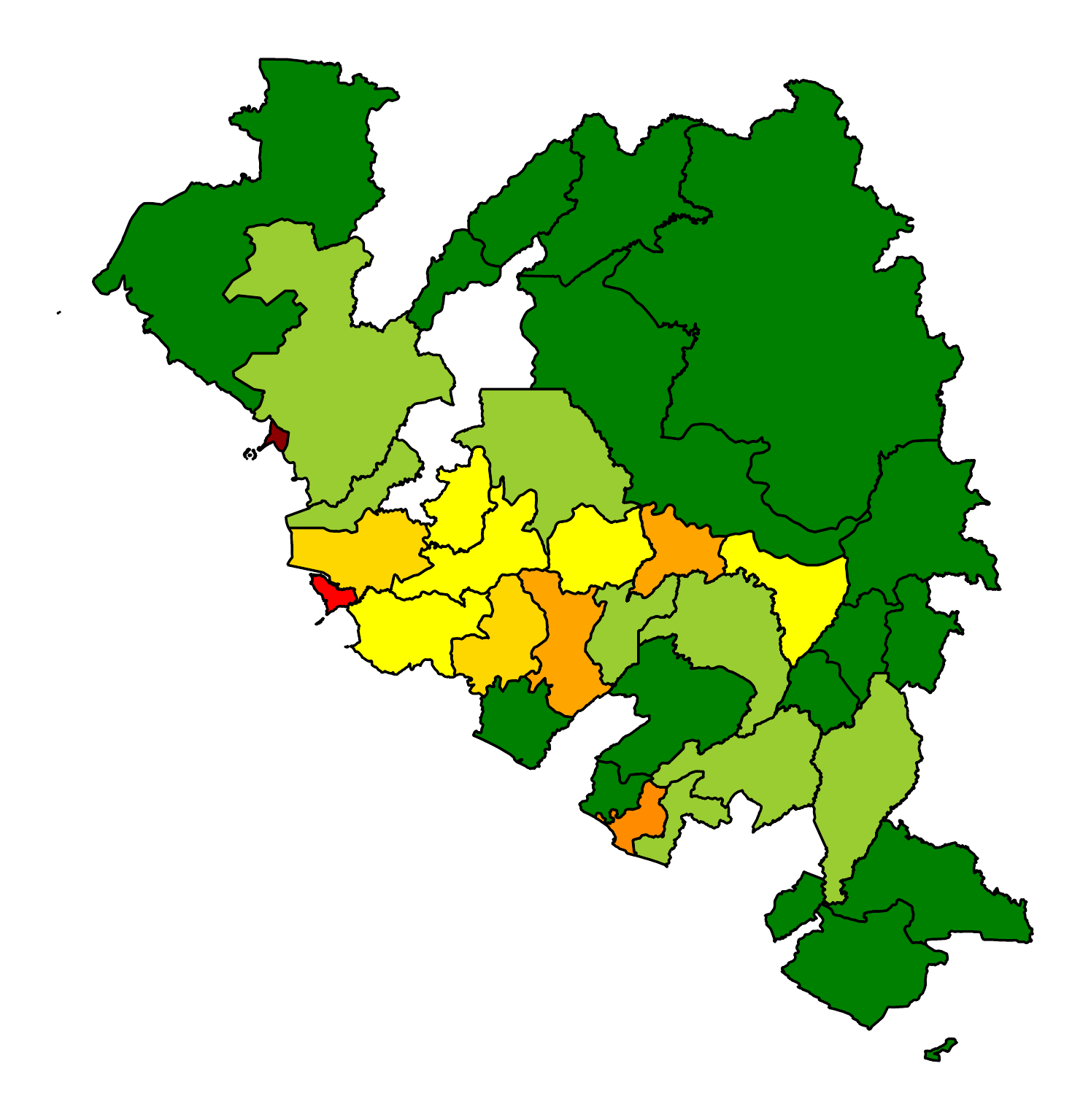}
		\caption*{Threshold = $900$}
	\end{subfigure}
	\caption{Geographical distribution of the overall cumulative number of infections per district the counterfactual 
	outbreaks where all individuals reduce their individual contacts by $25\%$ after the overall number of active infections 
	reaches a certain threshold. 
	Darker red (green) corresponds to high (low) number of cumulative infections.
	}
	\label{fig:geo-contact-reduction}
\end{figure}

To generate the contact network $\Gcal$ which underpins our model, we use a stochastic block model network whose parameters are informed by estimates of the reproduction 
numbers for each of the countries (Guinea, Liberia, and Sierra Leone) affected by the Ebola outbreak we focus on.
For computational reasons, for each of the $55$ districts where infection cases were identified, we create a set of nodes $\Vcal_i$ proportional to the population in the district, 
as estimated by the WorldPop project~\cite{lloyd2017high}, and assume $\cup_i \Vcal_i = \Vcal$ with $ |\Vcal| = \sum_{i} |\Vcal_i| = 8{,}000$ individuals.
Moreover, we add an edge between each pair of nodes $(u, v)$ nodes independently at random with probability
\begin{equation}
p(u, v) = 
\begin{cases}
10^{-2} & \text{if } (u, v) \text{ are in the same district,} \\
2.15 \cdot 10^{-3} & \text{if } (u, v) \text{ are in two contiguous districts in Guinea,} \\
3 \cdot 10^{-3} & \text{if } (u, v) \text{ are in two contiguous districts in Liberia,} \\
3.15 \cdot 10^{-3} & \text{if } (u, v) \text{ are in two contiguous districts in Sierra Leone,} \\
1.9 \cdot 10^{-3} & \text{if } (u, v) \text{ are in two contiguous districts in different countries,} \\
\end{cases}
\end{equation}
where we found contiguous districts using publicly available district-level shapefiles\footnote{\href{https://data.humdata.org/}{https://data.humdata.org/}} and set the values of the between-district 
probabilities using grid-search so that, at a country level, the basic reproduction number of the simulated outbreaks matches that estimated from real data (refer to Table~\ref{tab:r0-ebola} in Appendix~\ref{app:ebola}).
\begin{figure}[t]
		\centering
		\includegraphics[width=0.5\textwidth]{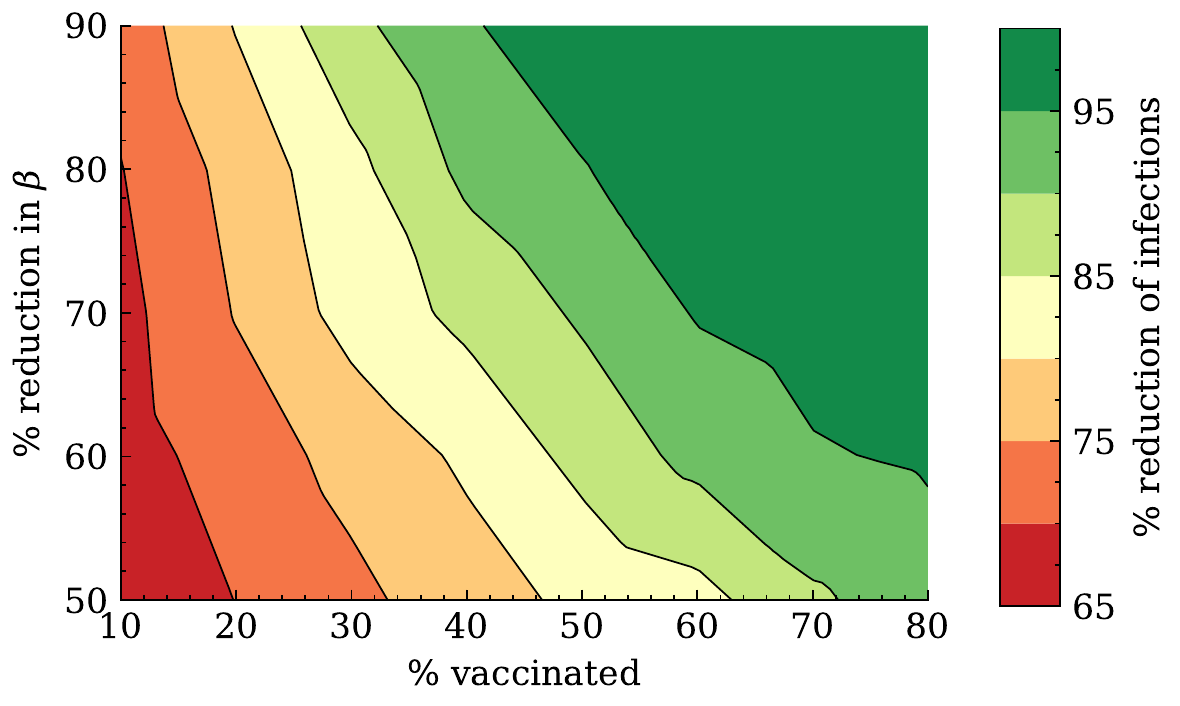}
		\caption{Effect of interventions where a percentage of the overall population receives a vaccine with a certain level of efficacy in a \emph{realistic} Ebola outbreak in West Africa. 
		The figure shows the average reduction in the cumulative number of infections under each intervention with respect to an outbreak sampled from the SIR model 
		defined by Eq.~\ref{eq:sir}.
		The disease specific and network parameters of the SIR model are calibrated using data from an Ebola outbreak in West Africa in 2013-2016~\cite{garske2017heterogeneities}
		and, for each level of vaccine adoption and efficacy, we generate $20$ counterfactual realizations using a variation of Algorithm~\ref{alg:counterfactual-hawkes} 
		(refer to Appendix~\ref{app:sir-counterfactuals}).}
		%
		\label{fig:vaccination}
\end{figure}

\begin{figure}[th]
	\begin{subfigure}{1\textwidth}
		\centering
		\includegraphics[width=0.7\linewidth]{FIG/MAP/legend2.pdf}	
	\end{subfigure}
	\begin{subfigure}{.33\textwidth}
		\centering
		\includegraphics[width=0.8\linewidth]{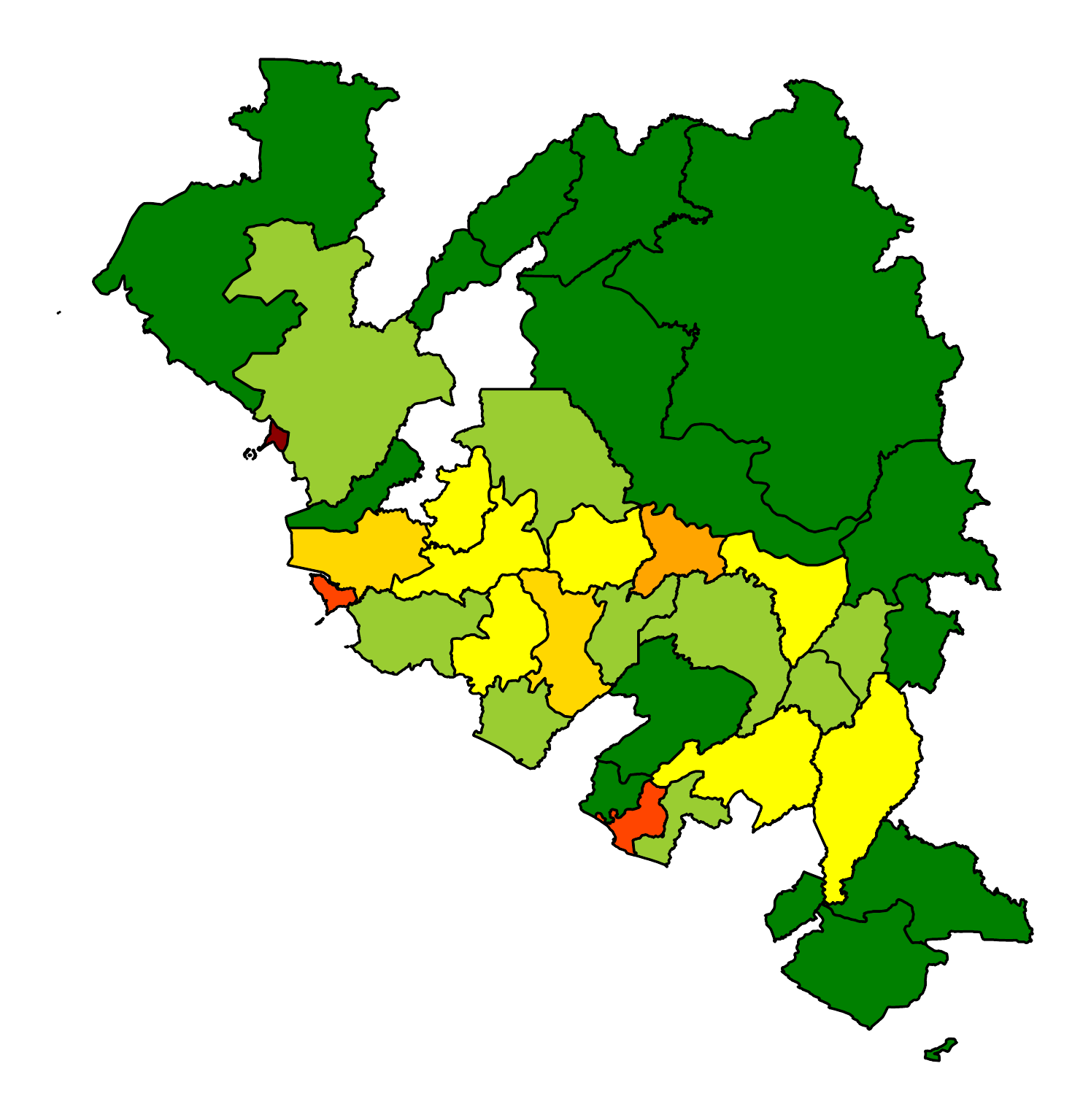}
		\caption*{$10\%$ vaccinated \\ $50\%$ reduction in $\beta$}
	\end{subfigure}%
	\begin{subfigure}{.33\textwidth}
		\centering
		\includegraphics[width=0.8\linewidth]{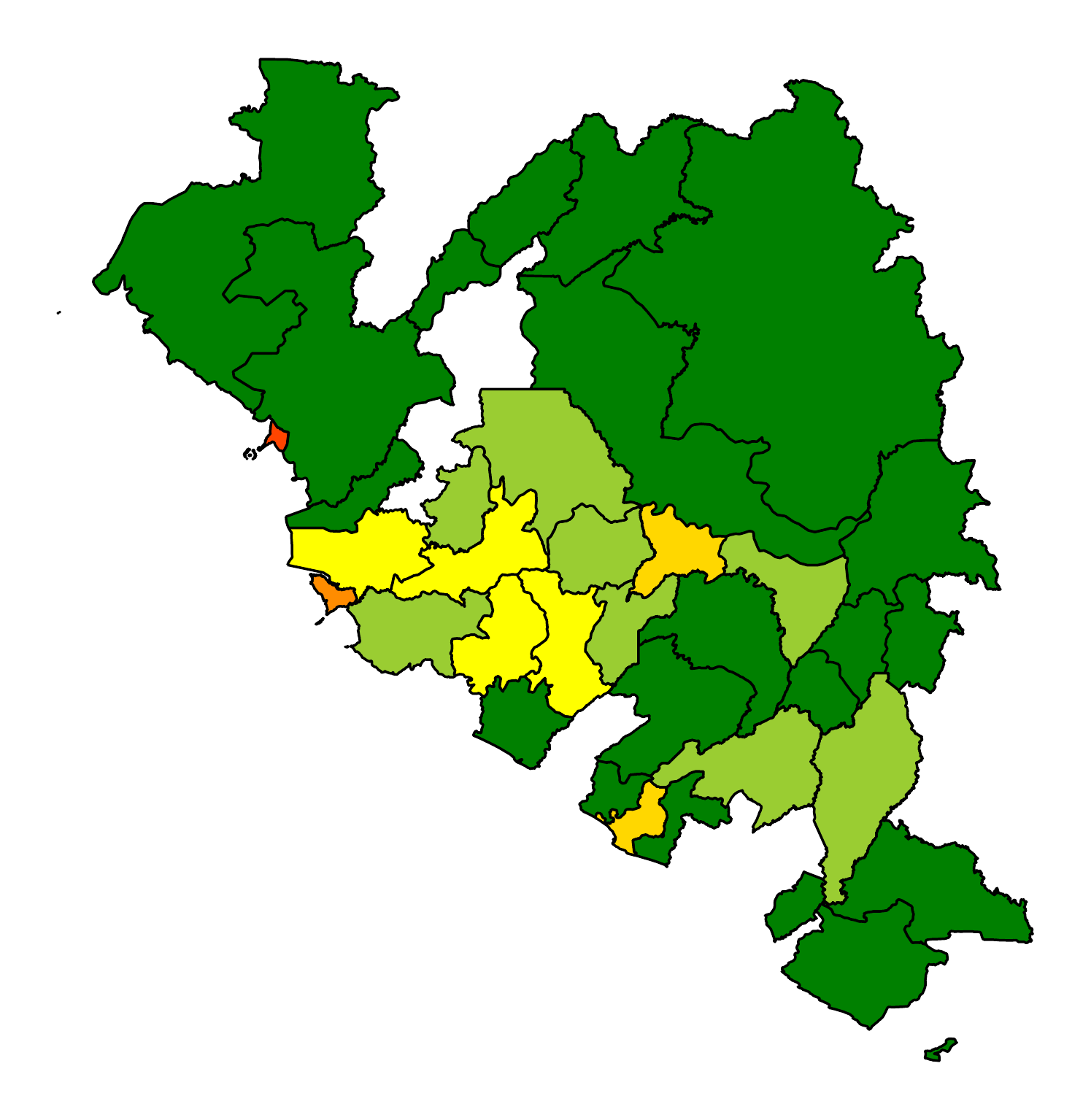}
		\caption*{$20\%$ vaccinated \\ $90\%$ reduction in $\beta$}
	\end{subfigure}%
	\begin{subfigure}{.33\textwidth}
		\centering
		\includegraphics[width=0.8\linewidth]{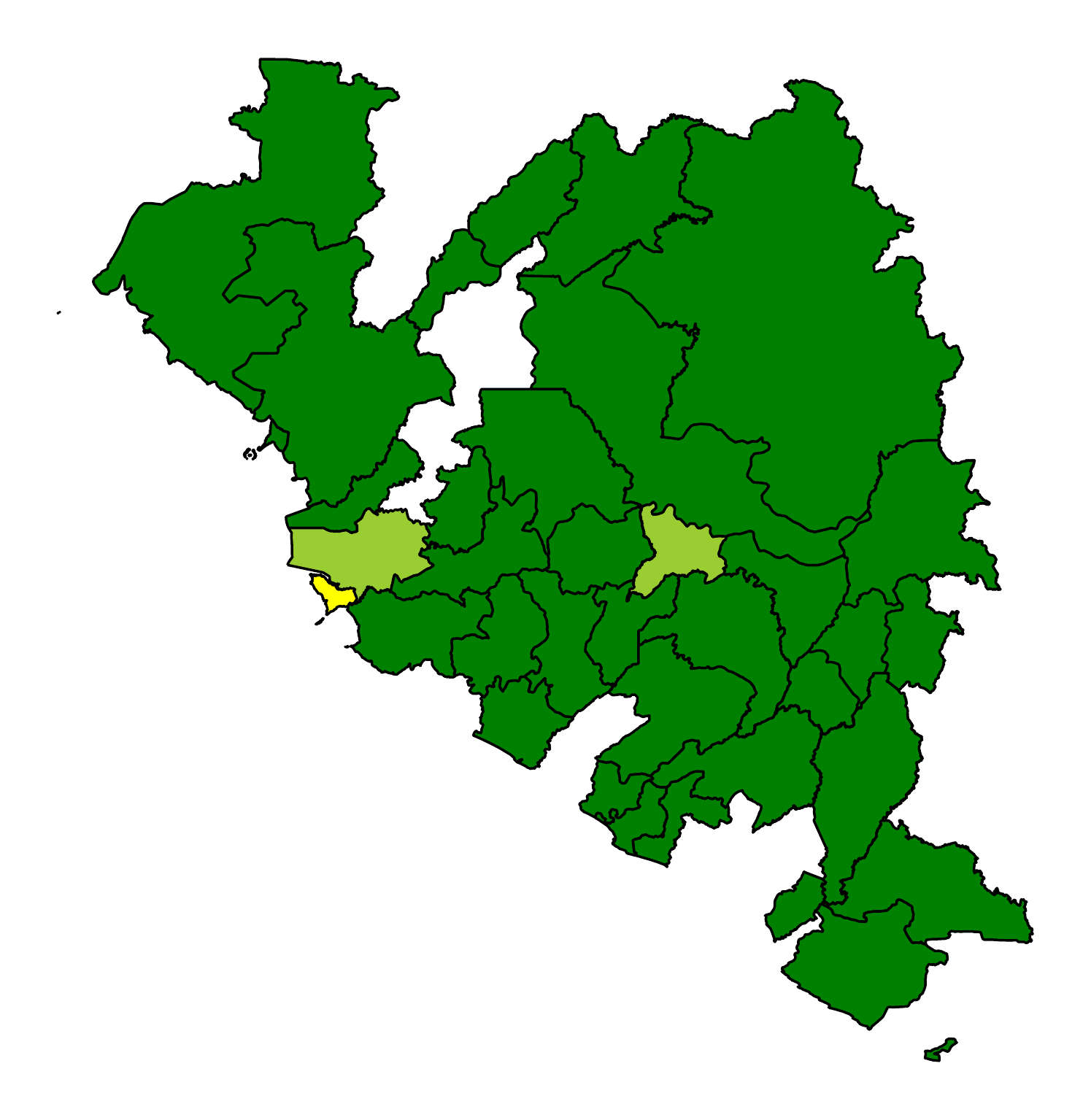}
		\caption*{$80\%$ vaccinated \\ $60\%$ reduction in $\beta$}
	\end{subfigure}
	\caption{Geographical distribution of the overall cumulative number of infections per district the counterfactual 
	outbreaks where percentage of the population receives a vaccine with a certain level of efficacy.
	Darker red (green) corresponds to high (low) number of cumulative infections.
	}
	\label{fig:geo-vaccine}
\end{figure}

In our experiments, we simulate a \emph{realistic} outbreak by sampling a realization from the above fitted model. To this end, for each real recorded case up to January 1, 2014~\cite{garske2017heterogeneities}, 
we sample a seed node at random from the same district as the observed case\footnote{The first real case was recorded on Dec 26, 2013. By January 1, 2014, there were six recorded cases
in four different districts.}. 
Figure~\ref{fig:geo-infections} shows the geographical distribution of seed infections and overall cumulative number of infections in the outbreak.
Then, given this sampled realization, we quantify the effect of two types of interventions by sampling counterfactual realizations using a variation of Algorithm~\ref{alg:counterfactual-hawkes} (refer 
to Appendix~\ref{app:sir-counterfactuals}):
\begin{enumerate}[leftmargin=0.8cm]
\item[---] \emph{Reduction of number of contacts}: individuals reduce their individual contacts after the overall number of active infections reaches a certain threshold. 
%
In one scenario, only individuals from the district with the highest incidence reduce their contacts within the district and get isolated from all other districts and, in another 
scenario, everyone reduces their individual contacts.

\item[---] \emph{Vaccination}: a percentage of the overall population receives a vaccine with a certain level of efficacy. Within our model, we measure
vaccine efficacy in terms of reduction of the value of the parameter $\beta$, which controls the infection rate between individuals.
\end{enumerate}
In each experiment, to estimate the average and confidence intervals of the outcome of interest (\eg, number of cases), we sample $20$ counterfactual realizations.

	
	
	

\xhdr{Results} Figures~\ref{fig:isolation}--\ref{fig:geo-contact-reduction} summarize the results for the interventions where individuals reduce their individual contacts after 
the overall number of active infections reaches a certain threshold.
Our results suggest that, at all threshold levels, reducing the individual contacts across all districts, even by just $5$\%, would have been more effective than isolating 
and reducing the contacts by $50$\% in the district with higher incidence.
Moreover, we also find that, for lower threshold values, the counterfactual outbreaks would have spread to fewer districts and the overall number of infections would have
been significantly lower.  
%
%

Figures~\ref{fig:vaccination}--\ref{fig:geo-vaccine} summarize the results for the interventions where a percentage of the overall population receives a vaccine with a certain level
of efficacy. Our results suggest that a high level of vaccine effectivity would have not been sufficient to reduce the number of infections if the percentage of the 
population who had received the vaccine was \emph{low}.
For example, if less than $20$\% of the population had received the vaccine, even a vaccine with a $90$\% effectivity would have been unable to reduce the infections 
by more than $70$\%. In contrast, if more than $80$\% of the population had received the vaccine, a vaccine with just a $60$\% effectivity would have reduced the 
infections by more than 95\%.
Moreover, we also find that, similarly as in the scenario where individuals reduce their individual contacts, the reductions in the overall number of infections would have 
also led to lower geographical dispersion. 

\xhdr{Remarks} 
%
%
We would like to acknowledge that, since counterfactual reasoning lies within level three in the ``ladder of causation''~\cite{pearl2009causality}, we cannot 
validate our counterfactual predictions using observational nor interventional experiments. 
That being said, we have made an intuitive assumption---monotonicity---about the causal mechanism of the world---our causal model of thinning---which specifies how changes
on the intensity function of a temporal point process may have lead to particular outcomes while holding ``every-thing else'' fixed, similarly as previous work~\cite{oberst2019counterfactual, tsirtsis2021counterfactual}. 
Moreover, we believe it may help domain experts (\eg, epidemiologists) perform counterfactual thinking, which has been argued to play a key role in human decision 
ma\-king~\citep{byrne2002mental, epstude2008functional}.

%% file: 070conclusions.tex
In this work, we have introduced a causal model of thinning for temporal point processes and shown that it satisfies a desirable 
counterfactual monotonicity condition that is sufficient to identify counterfactual dynamics in the process of thinning.
Building upon this causal model of thinning, we have developed a sampling algorithm to simulate counterfactual realizations of 
inhomogeneous Poisson processes and Hawkes processes.
Finally, we have evaluated our sampling algorithm using both synthetic and real epidemiological event data and shown that the 
counterfactual realizations our algorithm may give valuable insights to enhance targeted interventions.

Our work opens up many interesting avenues for future work. For example, our causal model of thinning, which builds upon the 
Gumbel-Max SCM and inherits the monotonicity assumption, may be realistic in some settings. 
However, it would be interesting to understand the sensitivity of counterfactual realizations to the specific choice of SCM. 
Moreover, a natural follow-up would be augmenting our model and algorithms to support temporal point processes with 
marks~\citep{schulam2017reliable}.
Finally, it would be important to carry out a user study in which the counterfactual realizations provided by our algorithm are 
shared with domain experts (\eg, epidemiologists) and evaluate our sampling algorithm using other real datasets from other 
applications.

%% file: 080appendix.tex
\section{Proof of invariance with respect to $\lambda_{\max}$} \label{app:invariance}
To prove that Algorithm~\ref{alg:counterfactual-lewis} is invariant with respect to the choice of $\lambda_{\max}$ as long as 
$\lambda_{\max} \geq \lambda_m(t)$ for all $m \in \Mcal$ and $t \in \RR^{+}$, we only need to prove that
\begin{equation*}
\text{(i)} \quad \lambda_{m}(t) P^{\Ccal \given X = 1, \Lambda = \lambda_{m}(t) \,;\, \text{do}(\Lambda = \lambda_{m'}(t))}(X) \quad \text{ whenever } \quad \lambda_m(t) < \lambda_{m'}(t)
\end{equation*} 
and
\begin{equation*}
\text{(ii)} \quad \lambda_{m}(t) + \left( \lambda_{max} - \lambda_{m}(t) \right) P^{\Ccal \given X = 0, \Lambda = \lambda_{m}(t) \,;\, \text{do}(\Lambda = \lambda_{m'}(t))}(X) \quad \text{ whenever } \quad \lambda_m(t) \geq \lambda_{m'}(t)
\end{equation*} 
are invariant to the specific choice of $\lambda_{max}$.

To prove that (i) is invariant, we first rewrite the counterfactual probability $P^{\Ccal \given X = 1, \Lambda = \lambda_{m}(t) \,;\, \text{do}(\Lambda = \lambda_{m'}(t))}$ in terms 
of uniform random variables, following Section~4.2 in Huijben et al.~\cite{huijben2022review}:
\begin{multline*}
	P^{\Ccal \given X = 1, \Lambda = \lambda_{m}(t) \,;\, \text{do}(\Lambda = \lambda_{m'}(t))}(X = 1) = \\
	\PP_{U_0, U_1} \left[ -\log\left(-\log(U_1)\right) + \log\left(\frac{\lambda_{m'}(t)}{\lambda_{\max}}\right) - \log\left(\frac{\lambda_{m}(t)}{\lambda_{\max}}\right) \right.
	> \\
	\left. -\log\left(-\log(U_1) - \frac{\log(U_0)}{1 - \frac{\lambda_{m}(t)}{\lambda_{\max}}}\right) + \log\left(1 - \frac{\lambda_{m'}(t)}{\lambda_{\max}}\right) - \log\left(1 - \frac{\lambda_{m}(t)}{\lambda_{\max}}\right) 
	 \right]
\end{multline*}
where $U_0, U_1 \sim U[0, 1]$. Now, for a fixed $U_0$ and $U_1$, the above inequality can be rewritten as follows:
\begin{align*}
-\log \left[ - \log(U_1) \frac{\lambda_{m}(t)}{\lambda_{\max}} \frac{\lambda_{\max}}{\lambda_{m'}(t)} \right] &> -\log \left[ \left( -\log(U_1) - \frac{\log(U_0)}{1-\frac{\lambda_m(t)}{\lambda_{\max}}} \right) \frac{1 - \frac{\lambda_{m}(t)}{\lambda_{\max}}}{1 - \frac{\lambda_{m'}(t)}{\lambda_{\max}}} \right] \\
- \log(U_1) \frac{\lambda_{m}(t)}{\lambda_{m'}(t)} \left( 1 - \frac{\lambda_{m'}(t)}{\lambda_{\max}} \right) &< -\log(U_1) \left(1 - \frac{\lambda_{m}(t)}{\lambda_{\max}}\right) - \log(U_0) \\
- \log(U_1) \left( \frac{\lambda_{m}(t)}{\lambda_{m'}(t)} - \frac{\lambda_{m}(t)}{\lambda_{\max}} \right) &< -\log(U_1) \left(1 - \frac{\lambda_{m}(t)}{\lambda_{\max}}\right) - \log(U_0)
\end{align*}
In the last inequality, the terms containing $\lambda_{\max}$ on the left and right hand side are identical and can be canceled. 
This proves that (i) is invariant to the specific choice of $\lambda_{max}$.

To prove that (ii) is also invariant, we proceed similarly as in (i) and first rewrite the counterfactual probability $P^{\Ccal \given X = 0, \Lambda = \lambda_{m}(t) \,;\, \text{do}(\Lambda = \lambda_{m'}(t))}$ in terms of uniform random variables:
\begin{multline*}
	P^{\Ccal \given X = 0, \Lambda = \lambda_{m}(t) \,;\, \text{do}(\Lambda = \lambda_{m'}(t))}(X = 1) = \\
	\PP_{U_0, U_1} \left[ -\log\left(-\log(U_0) - \frac{\log(U_1)}{\frac{\lambda_{m}(t)}{\lambda_{\max}}} \right) + \log\left(\frac{\lambda_{m'}(t)}{\lambda_{\max}}\right) - \log\left(\frac{\lambda_{m}(t)}{\lambda_{\max}}\right) \right.
	> \\
	\left. -\log\left(-\log(U_0) \right) + \log\left(1 - \frac{\lambda_{m'}(t)}{\lambda_{\max}}\right) - \log\left(1 - \frac{\lambda_{m}(t)}{\lambda_{\max}}\right) 
	 \right]
\end{multline*}
where $U_0, U_1 \sim U[0, 1]$. Now, for a fixed $U_0$ and $U_1$, the above inequality can be rewritten as follows:
\begin{equation*}
\log(U_0) \left( 1 - \frac{\lambda_m(t)}{\lambda_{m'}(t)} \right) \leq \log(U_1) \left( \frac{\lambda_{\max}}{\lambda_{m'}(t)} - 1 \right)
\end{equation*}
Now, using the fact that $-\log(U_0)$ and $-\log(U_1)$ are distributed as exponential random variables with rate $\lambda = 1$ 
and the CDF of the ratio $X = \log(U_0) / \log(U_1)$ of two exponential random variables with rate $\lambda_1$ is given by 
$\PP[X \leq x] = 1 / (1/x + 1)$, we have that:
\begin{equation*}
P^{\Ccal \given X = 0, \Lambda = \lambda_{m}(t) \,;\, \text{do}(\Lambda = \lambda_{m'}(t))}(X = 1) = \PP_{U_0, U_1} \left[ \frac{\log(U_0)}{\log(U_1)} > \frac{ \frac{\lambda_{\max}}{\lambda_{m'}(t)} - 1} {1 - \frac{\lambda_m(t)}{\lambda_{m'}(t)}} \right] = \frac{\lambda_{m'}(t) - \lambda_m(t)}{\lambda_{\max} - \lambda_m(t)}. 
\end{equation*}
Then, it readily follows that
\begin{equation*}
\left( \lambda_{\max} - \lambda_{m}(t) \right) \frac{\lambda_{m'}(t) - \lambda_m(t)}{\lambda_{\max} - \lambda_m(t)} = \lambda_{m'}(t) - \lambda_m(t).
\end{equation*}
This proves that (ii) is invariant to the specific choice of $\lambda_{max}$.

\section{Proof of Proposition~\ref{prop:monotonicity}} \label{app:prop:monotonicity}
If $\lambda_m(t_i) \geq \lambda_{m'}(t_i)$, then we have:
\begin{equation*}
\lambda_m(t_i) \geq \lambda_{m'}(t_i)
\implies\\ 
\lambda_m(t_i)\left(1-\frac{\lambda_{m'}(t_i)}{\lambda_{\text{max}}}\right) \geq \lambda_{m'}(t_i)\left(1-\frac{\lambda_m(t_i)}{\lambda_{\text{max}}}\right) 
\implies\\ 
\frac{1-\frac{\lambda_{m'}(t_i)}{\lambda_{\text{max}}}}{1-\frac{\lambda_m(t_i)}{\lambda_{\text{max}}}} 
\geq \frac{\frac{\lambda_{m'}(t_i)}{\lambda_{\text{max}}}}{\frac{\lambda_m(t_i)}{\lambda_{\text{max}}}},
\end{equation*}
%
Now, by Eq.~\ref{eq:distribution}, the last inequality is equivalent to: 
\begin{equation} \label{eq:stability}
\frac{P^{\Ccal \,;\, \text{do}(\Lambda = \lambda_{m'}(t_i))}(X = 0)}{P^{\Ccal \,;\, \text{do}(\Lambda = \lambda_m(t_i))}(X = 0)} \geq \frac{P^{\Ccal \,;\, \text{do}(\Lambda = \lambda_{m'}(t_i))}(X = 1)}{P^{\Ccal \,;\, \text{do}(\Lambda = \lambda_m(t_i))}(X = 1)},
\end{equation}
which is exactly the counterfactual stability property. 
Finally, by Theorem $2$ in~\cite{oberst2019counterfactual}, we know that the Gumbel-Max SCM satisfies the counterfactual stability 
property. As a result, Eq.~\ref{eq:stability} implies that 
\begin{equation*}
P^{\Ccal \given X = 0, \Lambda = \lambda_{m}(t_i) \,;\, \text{do}(\Lambda = \lambda_{m'}(t_i))}(X = 1) = 0, 
\end{equation*}
as desired. 
The proof for the second case is exactly the same.

\clearpage
\newpage

\section{Lewis'{} thinning algorithm} \label{app:lewis}
Within Algorithm~\ref{alg:lewis}, lines 4--6 sample an event from a Poisson
process with constant intensity $\lambda_{\text{max}}$ using inversion
sampling and lines 11--14 accept/reject the event according to the ratio
between the intensity of interest and $\lambda_{\text{max}}$ at the time
of the event.
 \IncMargin{1.2em}
\begin{algorithm}[h]
\small
\SetKwProg{Fn}{function}{:}{end}
  \textbf{Input}: $\lambda(t)$, $\lambda_{\text{max}}$, T. \\
  \textbf{Initialize}: $s = 0$, $\Hcal = \emptyset$. \\

\vspace{2mm}
\Fn{$\textsc{Lewis}(\lambda(t), \lambda_{\text{max}}, T)$} {
\While{\normalfont{true}}{
  $u_1 \sim \text{Uniform}(0, 1)$ \\
  $w \leftarrow - \ln u / \lambda_{\text{max}}$ \\
  $s \leftarrow s + w$ \\
  \If{$s > T$} { 
  	break
  }
  $\Hcal_{\max} \leftarrow \Hcal_{\max} \cup \{ s \}$ \\
  $u_2 \sim \text{Uniform}(0, 1)$ \\ 
  \If{$u_2 \leq \lambda(s) / \lambda_{\text{max}}$} {
  	$\Hcal \leftarrow \Hcal \cup \{ s \}$ \\
  }
}

\vspace{2mm}
\textbf{Return} $\Hcal$, $\Hcal_{\max} \backslash \Hcal$
}
\caption{Lewis'{} thinning algorithm} \label{alg:lewis}
\end{algorithm}
\DecMargin{1.2em}

\section{Thinning algorithm for Hawkes processes} \label{app:hawkes-superposition}
Algorithm~\ref{alg:hawkes-superposition} samples a sequence of events from a linear Hawkes process using its
branching process interpretation~\citep{moore2018maximum}, where $\textsc{Lewis}(\cdot)$ samples a sequence 
of events using Algorithm~\ref{alg:lewis}, $\gamma_0(t) = \mu$ and $\gamma_{i}(t) = \alpha g(t-t_i)$.
\IncMargin{1.2em}
\begin{algorithm}[h]
\small
\SetKwProg{Fn}{function}{:}{end}
\textbf{Input}: $\mu$, $\alpha$, $g(t)$, $\lambda_{\max}$, $T$. \\
\textbf{Initialize}: $\Hcal = \emptyset$. \\

\vspace{2mm}
\Fn{$\textsc{SampleHawkes}(\lambda(t), \lambda_{\text{max}}, T)$} {
$\Hcal \leftarrow \textsc{Lewis}(\gamma_{0}(t), \lambda_{\max}, T)$ \\

\vspace{2mm}
$\Hcal' \leftarrow \Hcal$ \\
\While{$| \Hcal' | > 0$} {
		$t_i \leftarrow \min_{t \in \Hcal'} t$ \\
		$\Hcal_{i}, \underline{\enskip} \leftarrow \textsc{Lewis}(\gamma_{i}(t), \lambda_{\max}, T)$ \\
		$\Hcal' \leftarrow \Hcal_{i} \cup \Hcal' \backslash \{ t_i \}$ \\
		$\Hcal \leftarrow \Hcal \cup \Hcal_{i}$ \\
}

\vspace{2mm}
\textbf{Return} $\Hcal$
}
\caption{It samples a sequence of events from a linear Hawkes process using its branching process interpretation.} \label{alg:hawkes-superposition}
\end{algorithm}
\DecMargin{1.2em}

\section{Additional Details about the Experiments on Real Data} \label{app:ebola}
Table~\ref{tab:r0-ebola} shows that, under the values of the between-district probabilities estimated using grid-search, the basic reproduction
numbers of the simulated outbreaks match, at a country level, those estimated from real data by WHO~\cite{who2014ebola}.

\begin{table}[thb]
	\centering
	\small
	\begin{tabular}{|c|c|c|}
		\hline
		Country           & $R_0$ (WHO)        & $R_0$ (simulated) \\ \hline
		Guinea (GN)       & 1.71 (1.44 -- 2.01) & 1.71 (1.66 -- 1.76)                        \\ \hline
		Liberia (LB)      & 1.83 (1.72 -- 1.94) & 1.83 (1.74 -- 1.91)                        \\ \hline
		Sierra Leone (SL) & 2.02 (1.79 -- 2.26) & 2.02 (1.95 -- 2.10)                        \\ \hline
	\end{tabular}
	\caption{Reproduction numbers ($R_0$) for each of the three countries affected by the Ebola outbreak estimated from real cases by WHO~\cite{who2014ebola} and 
	estimated from simulated cases. 
	In each cell, the first number is the average and the numbers in parentheses are the $95\%$ confidence interval. }
	\label{tab:r0-ebola}
\end{table}

\section{Sampling Counterfactual Events in a SIR Epidemic Model} \label{app:sir-counterfactuals}
As discussed in Section~\ref{sec:real}, the SIR model defined by Eq.~\ref{eq:sir} can be viewed as a (networked) multidimensional Hawkes process with 
stochastic triggering kernels defined by step functions. 
More specifically, let $t_i$ and $\tau_i = t_i + \Delta_i$ be the infection and recovery times of each node $i \in \Gcal$.
Then, the intensity $\EE[dY_i(t) \given \Hcal(t)]$ can be (re-)written as: 
\begin{equation*}
\EE[dY_i(t) \given \Hcal(t)] = \beta \sum_{j \in \Gcal(i)} g_i(t - t_j)
\end{equation*}
where $\Gcal(i)$ denotes the set of neighborhood of node $i$, $g_i(t) = [ 1 - Y_i(t) ] [ u(t) - u(t - \Delta) ]$ with $\Delta \sim \mathrm{Exp}(\delta)$ can be viewed as a stochastic triggering kernel, 
and $u(\cdot)$ is the step function.
As a result, we can adapt Algorithm~\ref{alg:counterfactual-hawkes}, originally developed for unidimensional linear Hawkes processes, to sample counterfactual realizations of the SIR model.
Algorithm~\ref{alg:counterfactual-sir} summarizes the resulting algorithm.

\IncMargin{1.2em}
\begin{algorithm*}[t]
	\small
	\textbf{Input}: $\beta_m$, $\delta$, $\beta_{m'}$, $\Gcal = (\Vcal, \Ecal)$, $\Gcal' = (\Vcal', \Ecal')$, $T$. \\
	\textbf{Initialize}: $\mathrm{queue} = \mathrm{PriorityQueue}(\{ i \given i \in \Vcal \,\mathrm{and}\, \mathrm{is\_seed(i)} \})$, $\beta_{max} = max(\beta_m, \beta_{m'})$, $\mathrm{processed} = \{\}$, $\mathrm{infector} = \{\}$.\\
	
	\vspace{2mm}
	\For{$i \in \Gcal$}{
		$\mathrm{processed}[i] = \mathrm{False}$\\
		\eIf{$\mathrm{is\_seed(i)}$}{
			$t_{ i} = 0$\\
			$\tau_{i} = t_{i} + \exp(\delta)$
		} {
			$t_{i} = \infty$ 
		}
	}
	\vspace{2mm}
	
	\While{$\neg \,\mathrm{queue.isEmpty()}$}{
		$i = \mathrm{queue}.pop()$\\
		\If{$\neg \, \mathrm{processed}[i]$}{
			$\mathrm{processed}[i] = \mathrm{True}$\\
			\For{$j \in \mathrm{\Gcal'(i)}$}{
				$\gamma_{m'}(t) = \beta_{m'} u(t - t_{i}) - \beta_{m'} u(t - \tau_{i})$\\
				\eIf{$\mathrm{infector}[j] == \mathrm{i}$}{
					$t = t_j$\\
					$\gamma_m(t) = \beta_m u(t - t_{i}) - \beta_m u(t - min(\tau_i, t))$\\
					$\Hcal_{m'} = \textsc{Cf}(\gamma_{m}(t), \gamma_{m'}(t), \{t\}, \beta_{max}, T)$\\
			\eIf{$\Hcal_{m'} \neq \emptyset$}{
				$t = \min_{t' \in \Hcal_{m'}} t'$
		}{
			$t = \infty$
		}
		}{
			$\Hcal, \_ = \textsc{Lewis}(\gamma_{m'}(t), \beta_{max}, T)$\\
			$t = \min_{t' \in \Hcal} t'$
		}
		
		\If{$t < t_j$}{
			$t_j = t$\\
			$\tau_j = t_j  + \exp(\delta)$\\
			$\mathrm{infector}[j] = i$\\
			$\mathrm{queue}.add(j, \mathrm{priority} = t_j)$
		}
		}
		}
	}
	
	\caption{It samples a counterfactual sequence of infections given a sequence of observed infections from the SIR process defined by Eq.~\ref{eq:sir}.}
	 \label{alg:counterfactual-sir}
\end{algorithm*}
\DecMargin{1.2em}